\documentclass[sigconf]{acmart}

\usepackage[colorinlistoftodos]{todonotes}
\usepackage{balance}
\usepackage{booktabs}
\usepackage{xspace} 
\usepackage{multirow}
\usepackage{diagbox}
\usepackage{subfigure}
\usepackage{braket}
\usepackage{bbm}
\usepackage[ruled]{algorithm2e}
\usepackage[skip=2pt]{caption}

\newcommand{\nop}[1]{}

\newcommand{\methodFont}{\textsl}
\newcommand{\ours}{\methodFont{DTIGNN}\xspace}
\newcommand{\oursNoR}{\methodFont{DTIGNN-I}\xspace}
\newcommand{\oursNoPR}{\methodFont{DTIGNN-I-T}\xspace}
\newcommand{\ourslayer}{{Attention based Spatial-Temporal GNNs Layer}\xspace}

\newcommand{\adj}{\mathcal{A}}
\newcommand{\graph}{\mathcal{G}}
\newcommand{\propmat}{\mathbf{\Gamma}}
\newcommand{\nt}{{Neural Transition Layer}\xspace}
\newcommand{\dynamic}{Phase-activated Adjacency Matrix\xspace}

\newcommand{\obt}{\textbf{x}_t}
\newcommand{\obtall}{\textbf{X}}
\newcommand{\oball}{\mathcal{X}}
\newcommand{\obtallz}{\textbf{Z}}
\newcommand{\mask}{\textbf{M}}

\newcommand{\DUni}{$D_{4\times 4}$\xspace}
\newcommand{\DHZ}{$D_{HZ}$\xspace}
\newcommand{\DNY}{$D_{NY}$\xspace}

\newtheorem{problem}{Problem}

\newcommand{\etc}{\textsl{etc.}\xspace}
\newcommand{\eg}{\textsl{e.g.}\xspace}

\AtBeginDocument{%
  \providecommand\BibTeX{{%
    \normalfont B\kern-0.5em{\scshape i\kern-0.25em b}\kern-0.8em\TeX}}}

\copyrightyear{2022}
\acmYear{2022}
\setcopyright{acmcopyright}
\acmConference[KDD '22]{Proceedings of the 28th ACM SIGKDD Conference on Knowledge Discovery and Data Mining}{August 14--18, 2022}{Washington, DC, USA}
\acmBooktitle{Proceedings of the 28th ACM SIGKDD Conference on Knowledge Discovery and Data Mining (KDD '22), August 14--18, 2022, Washington, DC, USA}
\acmPrice{15.00}
\acmDOI{10.1145/3534678.3539236}
\acmISBN{978-1-4503-9385-0/22/08}

\begin{document}
\begin{sloppypar}

\title{Modeling Network-level Traffic Flow Transitions on Sparse Data}

\author{Xiaoliang Lei}
\authornote{Both authors contributed equally to this research.}
\email{shawlenleo@stu.xjtu.edu.cn}
\affiliation{%
  \institution{Xi'an Jiaotong University}
   \city{Xi'an}
  \country{China}
}

\author{Hao Mei}
\authornotemark[1]
\email{hm467@njit.edu}
\affiliation{%
  \institution{New Jersey Institute of Technology}
  \city{Newark}
  \country{USA}
}

\author{Bin Shi}
\email{shibin@xjtu.edu.cn}
\affiliation{%
  \institution{Xi'an Jiaotong University}
  \city{Xi'an}
  \country{China}
}

\author{Hua Wei}
\email{hua.wei@njit.edu}
\authornote{Corresponding authors.}
\affiliation{%
 \institution{New Jersey Institute of Technology}
  \city{Newark}
  \country{USA}
 }
 
 \renewcommand{\shortauthors}{Xiaoliang Lei et al.}

\begin{abstract}
Modeling how network-level traffic flow changes in the urban environment is useful for decision-making in transportation, public safety and urban planning. The traffic flow system can be viewed as a dynamic process that transits between states (\eg, traffic volumes on each road segment) over time. 
In the real-world traffic system with traffic operation actions like traffic signal control or reversible lane changing, the system's state is influenced by both the historical states and the actions of traffic operations.
In this paper, we consider the problem of modeling network-level traffic flow under a real-world setting, where the available data is sparse (i.e., only part of the traffic system is observed).
We present \ours, an approach that can predict network-level traffic flows from sparse data. \ours models the traffic system as a dynamic graph influenced by traffic signals, learns the transition models grounded by fundamental transition equations from transportation, and predicts future traffic states with imputation in the process.
Through comprehensive experiments, we demonstrate that our method outperforms state-of-the-art methods and can better support decision-making in transportation.

\end{abstract}



\keywords{datasets, neural networks, gaze detection, text tagging}

\ccsdesc[500]{Information systems~Spatial-temporal systems}
\ccsdesc[300]{Computing methodologies~Neural networks}

\keywords{Traffic state modeling, traffic flow prediction, urban computing}

\maketitle

\section{Introduction}

Modeling how network-level traffic flow changes in the urban environment is useful for decision-making in various applications, ranging from transportation~\cite{lin2018efficient,fu2020compacteta}, public safety~\cite{wang2017no,zhang2017deep} to urban planning~\cite{jayarajah2018understanding}. For example, modeling the network-level traffic flow can help build a good simulator and serve as a foundation and a testbed for reinforcement learning (RL) on traffic signal control~\cite{wei2018intellilight}, routing~\cite{wang2019simple} and autonomous driving~\cite{wu2018stabilizing}.

The modeling of the traffic flow system can be viewed as a dynamic process that transits from state $s_t$ to state $s_{t+1}$. The state of the system can include traffic conditions like traffic volume, average speed, \etc  The ultimate objective to build a real predictor is to minimize the error between estimated state $\hat{s}_{t+1}$ and true state observation  $s_{t+1}$. Traditional transportation approaches assume the state transition model is given and calibrate parameters accordingly using observed data~\cite{kotsialos2002traffic, boel2006compositional,wang2022real,oh2018short}. However, the assumption on the state transition model is often far from the real world because the real-world system is a highly complex process, especially for the system in which humans are key players. 

Recently, growing research studies have developed data-driven methods to model state transitions. Unlike traditional transportation approaches, data-driven methods do not assume the underlying form of the transition model and can directly learn from the observed data. With the data-driven methods, a more sophisticated state transition model can be represented by a parameterized model like neural nets and provides a promising way to mimic the real-world state transitions. These data-driven models predict the next states based on the current state and historical states with 
spatial information, through deep neural networks like Recurrent Neural Network (RNN)~\cite{yao2018deep,yu2017spatio} or Graph Neural Network (GNN)
~\cite{kim2020stgrat,li2017diffusion,song2020spatial,wang2020traffic,wu2020connecting,zhang2020spatial,zhao2019t,guo2019attention,fang2021spatial}. As shown in Figure~\ref{fig:intro-phase}(a), these methods focus on directly minimizing the error between estimated state $s_{t+1}$ and true observed state $s_{t+1}$, with an end-to-end prediction model $P$. Although having shown great effectiveness over traditional methods, these approaches face two major challenges:

\begin{figure}[t!]
\centering
  \begin{tabular}{c}
  \includegraphics[width=.90\linewidth]{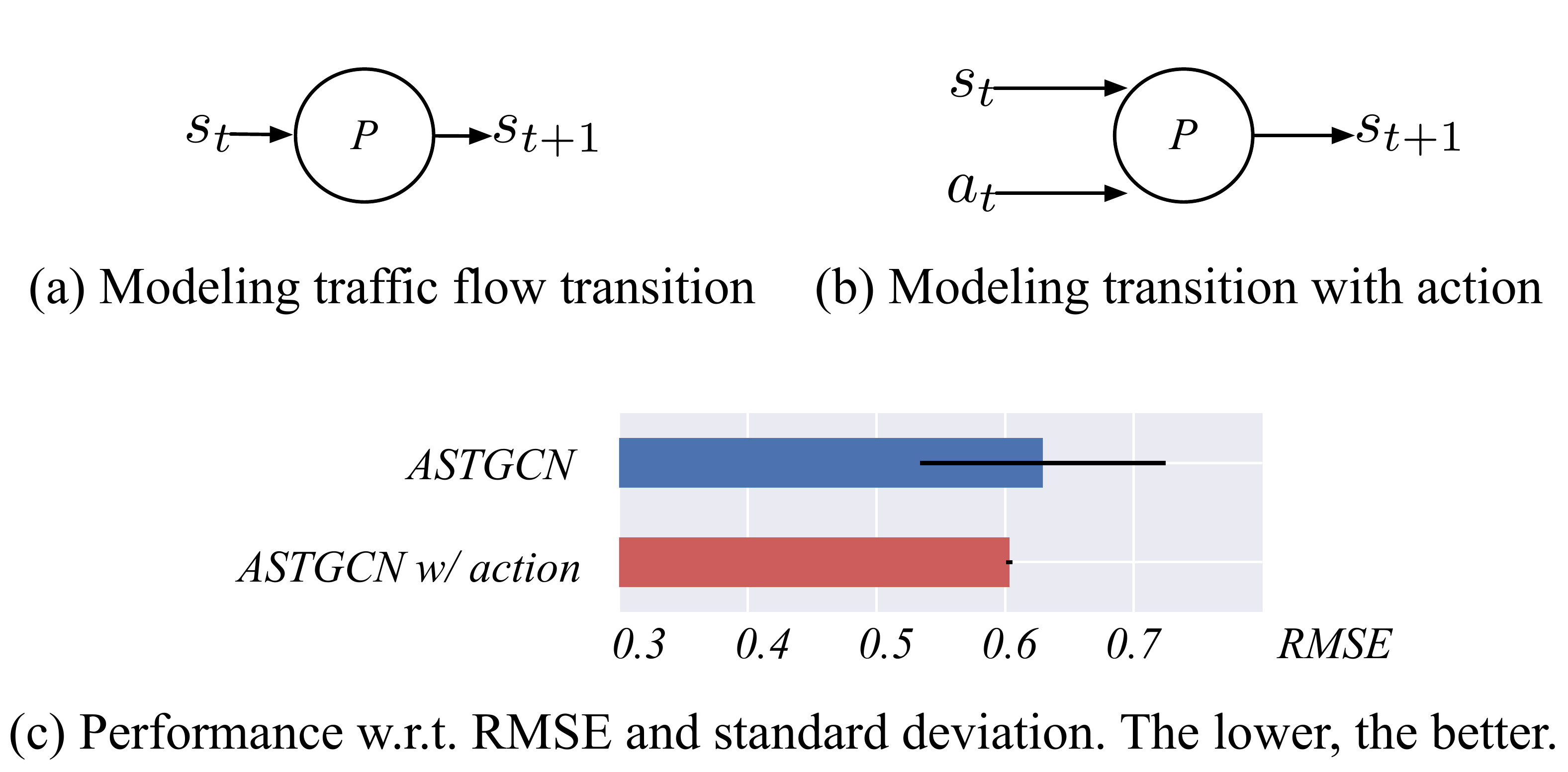}
  \end{tabular}
 \caption{Two perspectives in modeling traffic flow. (a) Existing data-driven models directly predict the next state $s_{t+1}$ based on its past states with a model $P$. (b) Modeling the traffic state transition takes both past states and traffic management actions. (c) The method with traffic management actions as input (ASTGCN~\cite{guo2019attention} w/ action) shows better performance.}
    \label{fig:intro-phase}
    \vspace{-5mm}
\end{figure}

\noindent$\bullet$~\textbf{The influence of traffic management actions.} In the real-world traffic system with traffic management actions like traffic signal control or reversible lane changing, the system's state is influenced not only by its previous states, but more importantly, by the actions of traffic management actions. Simply taking previous states (\eg., volumes) as input may cause conflicting learning problems. For example, given a road segment $A$ with $K$ vehicles at time $t$, the road segment has traffic signals at its end, controlling whether the vehicles can leave the road or not. When $A$ is having green light, the future traffic volume on the road is likely to drop, but if $A$ is having red light, the future traffic volume is likely to rise. When the model only considers historical traffic volumes, the conflicting traffic volume will confuse the learning process. As a result, the learned model is likely to predict $s_{t+1}$ with a large error or variance. As is shown in Figure~\ref{fig:intro-phase}, if we take the traffic management actions (\eg, how traffic signal changes) into consideration, the traffic flow will be predicted more accurately. To the best of our knowledge, none of the existing literature has integrated traffic actions into data-driven models.

It is worth noting that modeling the traffic flow state transition with traffic management actions is more than improving the prediction accuracy. A well-trained state transition model with traffic management actions can be utilized to provide actionable insights: it can be used to find the best decision to mitigate the problems in traffic flow system (e.g., traffic congestion), and then on prescribing the best actions to implement such a decision in the physical world and study the impact of such implementation on the physical world.

\noindent$\bullet$~\textbf{The sparsity of traffic data.} In most real-world cases, the available observation is sparse, i.e., the traffic flow states at every location are difficult to observe. It is infeasible to install sensors for every vehicle in the road network or to install cameras covering every location in the road network to capture the whole traffic situation. Most real-world cases are that the camera data usually only covers some intersections of the city, and the GPS trajectories may only be available on some cars, like taxis. As data sparsity is considered as a critical issue for unsatisfactory accuracy in machine learning, directly using datasets with missing observations to learn the traffic flow transitions could make the model fail to learn the traffic situations at the unobserved roads. 

To deal with sparse observations, a typical approach is to infer the missing observations first~\cite{garcia2010pattern,batista2002study,li2019misgan, luo2018multivariate, yoon2018gain, qin2021network} and then learn the model with the transition of traffic states. This two-step approach has an obvious weakness, especially in the problem of learning transition models with some observations entirely missing. For example, mean imputation~\cite{garcia2010pattern} is often used to infer the missing states on the road by averaging the states from nearby observed roads. However, not all the traffic from nearby roads would influence the unobserved road because of traffic signals, making the imputed traffic states different from the true states. Then training models on the inaccurate data would further lead to inaccurate predictions. A better approach is to integrate imputation with prediction because they should inherently be the one model: the traffic state on the unobserved road at time $t$ is actually influenced by the traffic flows before $t$, including the flows traversed from nearby roads and the remaining flows of its own, which is also unobserved and needs inference.

In this paper, we present \ours, a GNN-based approach that can predict network-level traffic flows from sparse observations, \vspace{-1mm}with \underline{\textbf{D}}ynamic adjacency matrix, \underline{\textbf{T}}ransition equations from transportation, and \underline{\textbf{I}}mputation. 
To model the influence of traffic management actions, \ours represents the road network as a dynamic graph, with the road segments as the nodes, road connectivity as the edges, and traffic signals changing the road connectivity from time to time. To deal with the sparse observation issue, we design a Neural Transition Layer to incorporate the fundamental transition equations from transportation, with theoretical proof on the equivalence between Neural Transition Layer and the transportation equations. \ours further imputes the unobserved states iteratively and predict future states in one model. The intuition behind is, the imputation and prediction provided by data-driven models should also follow the transition model, even though some flows are unobservable. 

We conduct comprehensive experiments using both synthetic and real-world data. We demonstrate that our proposed method outperforms state-of-the-art methods and can also be easily integrated with existing GNN-based methods. The ablation studies show that the dynamic graph is necessary, and integrating transition equation leads to an efficient learning process. We further discuss several interesting results to show that our method can help downstream decision-making tasks like traffic signal control.
\raggedbottom

\section{Related Work}
\label{sec:related}

\begin{figure*}[t!]
\centering
\vspace{-3mm}
  \includegraphics[width=.85\linewidth]{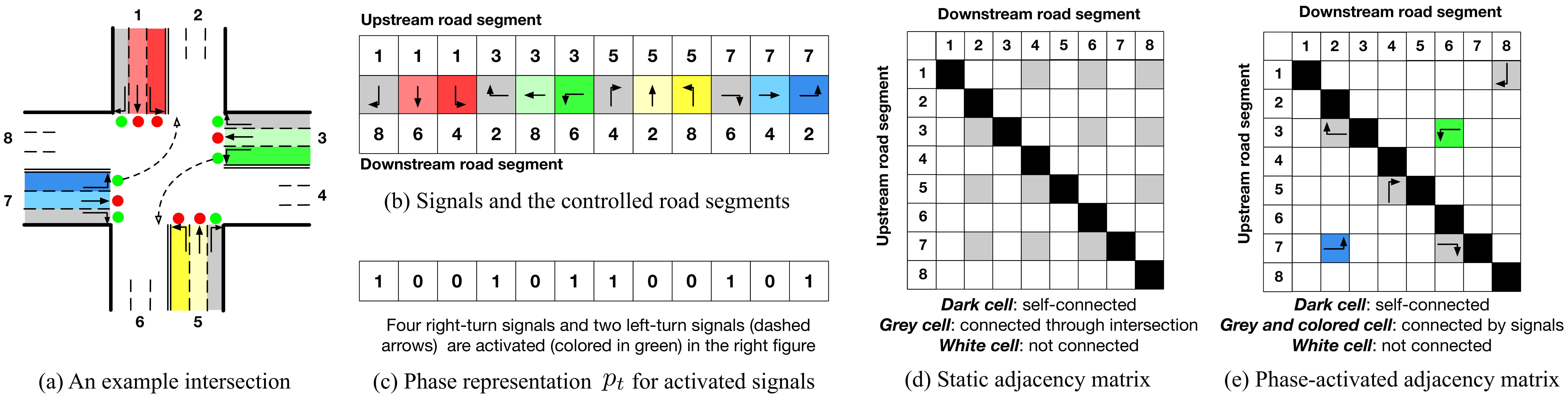}
 \caption{Illustration of traffic signals and their influence on the connectivity between road segments. (a) An intersection with eight connecting road segments and traffic signals. (b) The signals set and their controlled road segments in the example, with right-turn signals illustrated in grey cells. (c) The phase representation for the activated signals in the example intersection.(d) The static adjacency matrix between road segments induced by road network structure. (e) the dynamic adjacency matrix induced by traffic signals. Best viewed in color.}
 \vspace{-5mm}
    \label{fig:prelim}
\end{figure*}

\textbf{Traffic Prediction with Deep Learning.}
Recently, many efforts have been devoted to developing traffic prediction techniques based on various neural network architectures. One straightforward solution is to apply the RNN, or convolutional networks (CNN) to encode the temporal~\cite{yu2017deep, liu2016predicting} and spatial dependency~\cite{zhang2017deep, yao2018deep,yao2019revisiting}. Recent works introduce GNN to learn the traffic networks~\cite{zhang2020spatial}. DCRNN~\cite{li2017diffusion} utilizes the bi-directional random walks on the traffic graph to model spatial information and captures temporal dynamics by RNN. Transformer models~\cite{wang2020traffic,kim2020stgrat,wu2020connecting,zhao2019t,zhang2020traffic} utilize spatial and temporal attention modules in transformer for spatial-temporal modeling. STGCN~\cite{yu2017spatio} and GraphWaveNet~\cite{wu2019graph} model the spatial and temporal dependency separately with graph convolution and 1-D convolution. Later studies~\cite{song2020spatial,guo2019attention} attempt to incorporate spatial and temporal blocks altogether by localized spatial-temporal synchronous graph convolution module regardless of global mutual effect. However, when predicting traffic states, all the previous GNN-based models assume that the node feature information is complete, which is unfeasible in real-world cases. Our work can be easily integrated into these GNN-based methods and handle incomplete and missing feature scenarios, which have not been well explored in existing solutions.
~\noindent\\~\textbf{Traffic Inference with Missing Data.}
Incomplete and missing data is common in real-world scenario. In machine learning area, imputation techniques are widely used for data completion, such as mean imputation~\cite{garcia2010pattern},  matrix factorization~\cite{koren2009matrix,mazumder2010spectral}, KNN~\cite{batista2002study} and generative adversarial networks~\cite{li2019misgan, luo2018multivariate, yoon2018gain}. However, general imputation methods are not always competent to handle the specific challenge of spatial and temporal dependencies between traffic flows, especially when traffic flows on unmonitored road segments are entirely missing in our problem setting.

Many recent studies~\cite{tang2019joint,yi2019citytraffic} use graph embeddings or GNN to model the spatio-temporal dependencies in data for network-wide traffic state estimation. Researchers further combine temporal GCN with variational autoencoder and generative adversarial network to impute network-wide traffic state~\cite{qin2021network}. However, most of these methods assume the connectivity between road segments is static, whereas the connectivity changes dynamically with traffic signals in the real world. Moreover, all the graph-based methods do not explicitly consider the flow transition between road segments in their model. As a result, the model would likely infer a volume for a road segment when the nearby road segments do not hold that volume in the past.



\section{Preliminaries}
\label{sec:prelim}

\begin{definition} [Road network and traffic flow]
A road network consists of road segments, intersections, and traffic signals. Road segments are connected through intersections, and their connectivity changes with the action of traffic signals. The traffic flow on a road segment is directional, with incoming traffic flow from several upstream road segments, and different outgoing traffic flows to downstream road segments.
\end{definition}

\vspace{-3mm}
\begin{definition}[Traffic signal action $p$]
The traffic signal action $p_t \in \{0, 1\}^{P}$ represents a traffic signal phase~\cite{wei2019presslight} at time $t$, where there are $P$ signals in total, and its $k$-th value $p_t[k]=1$, indicating the signal is green, and corresponding traffic flows are allowed to travel from upstream roads to downstream roads.
\end{definition}

\vspace{-3mm}
\begin{definition}[Road adjacency graph $\graph$]
We represent the road network as a directed dynamic graph $\graph_t=\{\mathcal {R}, \adj_t\}$ at time $t$, where $\mathcal {R}=\{r^1,...,r^N\}$ is a set of $N$ road segments and $\adj_t \in \mathbb{R}^{N \times N}$ is the adjacency matrix indicating the connectivity between road segments at time t. In this paper, $\adj_t$ is dynamically changing with time $t$ due to the actions of traffic signals. 
\end{definition}

\vspace{-1mm}
Figure~\ref{fig:prelim}(a) shows an example intersection. Figure~\ref{fig:prelim}(b) indicates the intersection has 12 different signals, each of which changes the traffic connectivity from the upstream road segments to corresponding the downstream roads. Now the intersection is at signal phase $p_t$. In Figure~\ref{fig:prelim}(c), $p_t[0]=1$ denotes that traffic flows can move from the upstream road $\#1$ to downstream road $\#8$.
A static adjacency matrix is constructed based on the road network topology, and a dynamic adjacency matrix can be constructed based on the connectivity of between upstream and downstream road segments with signal phases. 




\vspace{-1mm}
\begin{definition}[Observability mask $\mask$]
In real-world traffic systems, some road segments are unobserved for the entire time due to the lack of sensors. We denote the observability of traffic states as a static binary mask matrix $\mask \in \{0,1\}^{N \times F}$, and $F$ is the length of a state feature (also called channels),  where $F=3$ when the road has three outgoing flows. When the road segment is observed, $\mask^i = \textbf{1}^{F}$; when the road segment is unobserved, $\mask^i = \textbf{0}^{F}$.
\end{definition}

\vspace{-3mm}
\begin{definition}[Graph state tensor $\oball$, observed $\dot{\oball}$, unobserved $\ddot{\oball}$, predicted $\widehat{\oball}$, merged $\oball'$]
We use $\obt^i \in  \mathbb{R}^F$ to denote the traffic volumes on road $r^i \in \mathcal {R}$ at time $t$. $\obtall_t=(\obt^1, \obt^2, \cdots, \obt^N) \in \mathbb{R}^{N \times F}$ denotes the states of all the road segments at time $t$, including the observed part $\dot{\obtall}_t=\obtall_t \odot \mask$ and unobserved part $\ddot{\obtall}_t=\obtall_t \odot (\textbf{1}-\mask)$, where $\odot$ is the element-wise multiplication. The graph state tensor $\oball=(\obtall_1, \obtall_2, \cdots, \obtall_T) \in \mathbb{R}^{N \times F \times T} $ denotes the states of all the road segments of all time, with $\dot{\oball}=(\dot{\obtall}_1, \dot{\obtall}_2, \cdots, \dot{\obtall}_T)$ as observed state tensor and $\ddot{\oball}$ as unobserved state tensor. The predicted values of $\oball$ will be denoted with $\widehat{\oball}$, and the merged values for $\oball$ will be denoted as $\oball'$.
\end{definition}


\begin{figure*}[t!]
\centering
  \includegraphics[width=.90\linewidth]{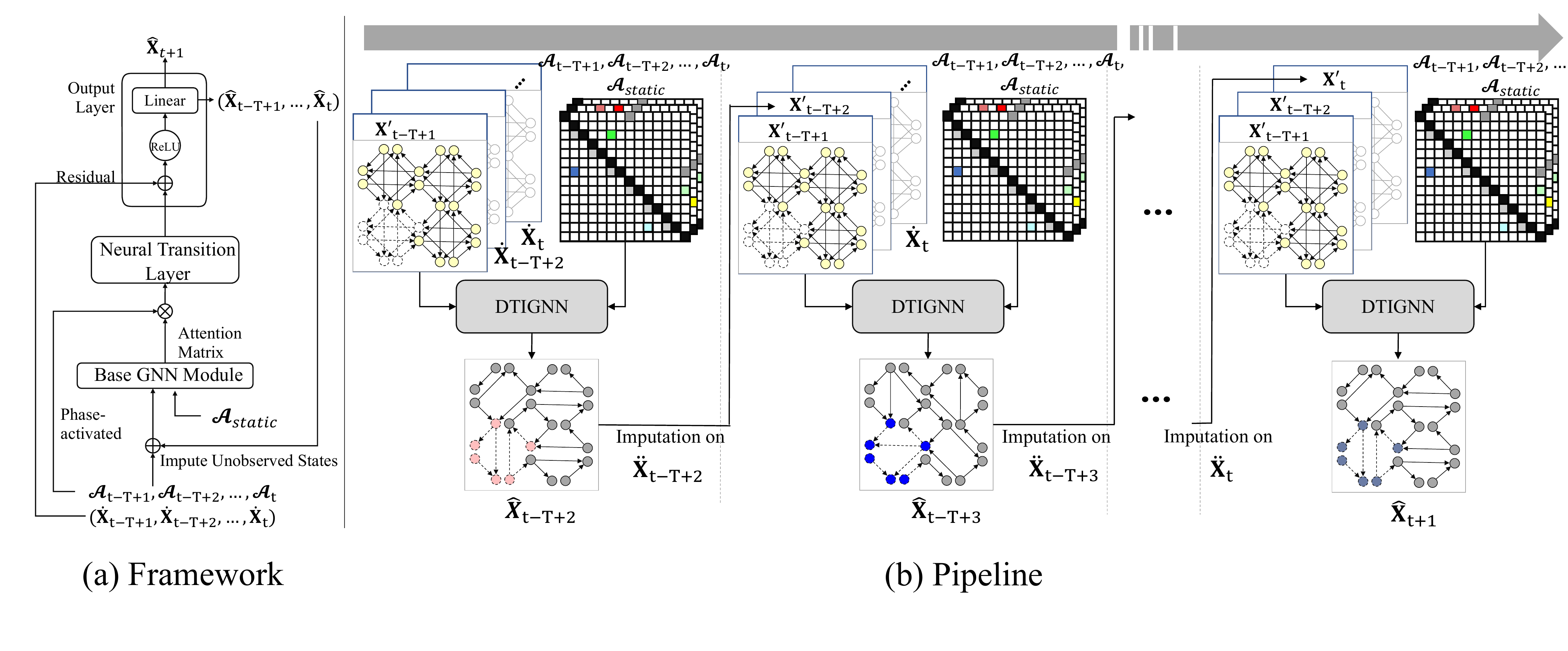}
      \vspace{-5mm}
 \caption{Model framework and training pipeline. (a) The framework of \ours network. The network can be built upon exiting spatial-temporal GNNs with our \ourslayer appended after existing GNNs. (b) The training pipeline of \ours. One training round goes through data in an auto-regressive way, and the predictions from previous time steps are used as imputations to update the unobserved data at the current time step.}
    \label{fig:framework}
      \vspace{-5mm}
\end{figure*}

\vspace{-3mm}
\begin{problem}[Traffic Flow Transition Modeling]
Based on the definitions above, the modeling of traffic flow transition is formulated as: given the observed tensor $\dot{\oball}$ on a road adjacency graph $\graph$, the goal of modeling traffic flow transition is to learn a mapping function $f$ from the historical $T$ observations to predict the next traffic state,
\begin{equation}
\setlength{\abovedisplayskip}{-2pt}
    [ \dot{\obtall}_{t-T+1}, \cdots, \dot{\obtall}_{t}; \graph_{t-T+1}, \cdots, \graph_{t} ]  \overset{f} \longrightarrow  [ \obtall_{t+1}]
\setlength{\belowdisplayskip}{-2pt}
\end{equation}
\end{problem}

It's worth noting that the modeling of state transitions focuses on predicting the next state $\obtall_{t+1}$, rather than predicting future $T'$ steps ($T'\geq 2$), since it requires future $\graph_{t+T'-1}$ which is influenced by future traffic operation actions. The learned mapping function $f$ can be easily extended to predict for longer time steps in an auto-regressive way if future traffic operation actions are preset.

\section{Methodology}
\vspace{-1mm}
\subsection{Overall Framework}
Figure~\ref{fig:framework} shows the framework of our proposed network which mainly consists of three components: a base GNN module, a Neural Transition Layer, and an output Layer. The base GNN module takes the static adjacency matrix $\adj_{static}$ and observed state tensor $\dot{\oball}=(\dot{\obtall}_{t-T+1}, \dot{\obtall}_{t-T+2}, \cdots, \dot{\obtall}_t)$ as input, and output an attention matrix $Att$. Grounded by the transition equations from transportation, the \nt takes the $\propmat$ along with dynamic \dynamic $\adj$ and $\dot{\oball}$ to predict the state tensor $\widehat{\obtall}_{t+1}$. As is shown in Figure~\ref{fig:framework}(b), there are iterative imputation steps from the time step $t-T+1$ toward $t+1$, where the unobserved part of the predict state tensor $\widehat{\ddot{\obtall}}_{t-\tau+1}$ from the time step $t-\tau$ would be merged 
with $\dot{\obtall}_{t-\tau+1}$ and used for the prediction for $\dot{\obtall}_{t-\tau+2}$. 
Then the Output Layer generates a prediction of $\oball_{t+1}$.
The details and implementation of the models will be described in the following section.

\vspace{-1mm}
\subsection{\dynamic}
As introduced in Section~\ref{sec:related}, existing traffic flow prediction methods rely heavily on a static adjacency matrix $\adj_{static}$ generated from the road network, which is defined as:
\begin{equation}
\small
    \adj_{static}^{i,j}=\begin{cases}
1, & \text{road segment $i$ is the upstream of $j$} \\[-1ex]
0, & \text{otherwise} \\
\end{cases}
\end{equation}

Considering that traffic signal phases change the connectivity of the road segments, we construct a \dynamic $\adj_t$ through the signal phase $p_t$ at time $t$:
\begin{equation}
\small
    \adj_t^{i,j}=\begin{cases}
1, & \text{${\exists}$ $k\in \{1, \cdots,P\}$, $up[k]=i$, $down[k]=j$, $p_t[k]=1$} \\[-1ex]
0, & \text{\ otherwise}
\end{cases}
\end{equation}
where $p_t[k]$ denotes the activation state of the $k$-th element in $p_t$ at time $t$, $up[k]$ and $down[k]$ denote the upstream road segment and the downstream road segment that associated with the $k$-th signal. Intuitively, upstream road $i$ will be connected with road $j$, when the $k$-th signal in current phase $p_t$ is green.



\vspace{-1mm}
\subsection{Transition-based Spatial Temporal GNN}

\subsubsection{Base GNN Module}

To capture the complex and spatial-temporal relationships simultaneously, a base GNN module is applied. GNNs update embeddings of nodes through a neighborhood aggregation scheme, where the computation of node representation is carried out by sampling and aggregating features of neighboring nodes. In this paper, any graph attention network (GAT) or graph convolutional network (GCN) model with multiple stacked layers could be adopted as our base GNN module. Without losing generality, here we introduce the first layer for GAT and GCN model stacks and then how to integrate them into our model.
~\noindent\\$\bullet$~The classic form of modeling dependencies in GAT can be formulated as~\cite{guo2019attention}: $\textbf{S}=\textbf{V}_s \cdot \sigma((\mathcal{X}\textbf{W}_s^{(1)})\textbf{W}_s^{(2)}(\textbf{W}_s^{(3)}\mathcal{X})^\intercal +\textbf{b}_s)$, 
where $\mathcal{X}=(\obtall_1,\obtall_2,\cdots,\obtall_T) \in \mathbb{R}^{N \times F\times T}$ is the input of the GAT module, $F$ is the channels of input feature. $\textbf{V}_s\in \mathbb{R}^{N \times N}$,$\textbf{b}_s \in \mathbb{R}^{N \times N}$,$\textbf{W}_s^{(1)}$
$\in \mathbb{R}^T$,$\textbf{W}_s^{(2)} \in \mathbb{R}^{F\times T}$, and $\textbf{W}_s^{(3)} \in \mathbb{R}^F$ are learnable parameters. $\sigma$ denotes sigmoid activation function. Then the attention matrix $\textbf{Att}\in\mathbb{R}^{N\times N}$ is calculated as:
\begin{equation}
\small
\label{eq:att}
\begin{aligned}
    \textbf{Att}^{i,j}=\frac{exp(\textbf{S}^{i,j})}{\sum^N_{j=1}exp(\textbf{S}^{i,j})}
\end{aligned}
\end{equation}
where the attention coefficient is calculated by softmax function to capture the influences between nodes.
~\noindent\\$\bullet$~The classic form of modeling spatial influences in GCN can be formulated as $\textbf{H}^{(1)}=\sigma(\widehat{\textbf{A}}\mathcal{X}\textbf{W})$,
where $\textbf{H}^{(1)} \in \mathbb{R}^{N\times F\times T}$ denotes the output of the 1st layer in GCN module, $\textbf{W} \in \mathbb{R}^{F \times F}$ is a learnable parameter matrix.
~$\sigma$ represents nonlinear activation. $\widehat{\textbf{A}}=\textbf{D}^{-\frac{1}{2}}\adj_{static} \textbf{D}^{-\frac{1}{2}} \in \mathbb{R}^{N\times N}$ represents the normalized adjacency matrix, where $\adj_{static}$ is the static adjacency matrix, and $\textbf{D}_{ii}=\sum_j \adj_{static}^{i,j}$ is the diagonal matrix.

The output of GCN model is then feed into the following equation to align the outputs with GAT-based module: 
$\textbf{Q}= \textbf{V}_q \cdot \sigma((\textbf{H}^{(l)}\textbf{W}_q^{(1)})$
$\textbf{W}_q^{(2)}\ (\textbf{W}_q^{(3)}\textbf{H}^{(l)})^\intercal +\textbf{b}_q)$, and we have the attention matrix for GCN-based module:
\begin{equation}
\small
\begin{aligned}
    \textbf{Att}^{i,j}=\frac{exp(\textbf{Q}^{i,j})}{\sum^N_{j=1}exp(\textbf{Q}^{i,j})}
\end{aligned}
\end{equation}
where $\textbf{H}^{(l)} \in \mathbb{R}^{N\times F\times T}$ denotes the output of the last layer ($l$-th) in the GCN model, $\textbf{V}_q\in \mathbb{R}^{N \times N}$,$\textbf{b}_q \in \mathbb{R}^{N \times N}$,$\textbf{W}_q^{(1)}$
$\in \mathbb{R}^T$,$\textbf{W}_q^{(2)} \in \mathbb{R}^{F\times T}$, and $\textbf{W}_q^{(3)} \in \mathbb{R}^F$ are learnable parameters. $\sigma$ denotes sigmoid activation function.

In this case, the final outputs of both GAT- and GCN-based modules are attention matrix $\textbf{Att}$, which will be used in later parts of our model to facilitate imputation on subsequent layers in \ours.

\vspace{-1mm}
\subsubsection{\nt}

After the base GNN module, we can obtain the attention matrix $\textbf{Att}$. Then a dot product operation between attention matrix and \dynamic is required to get activated proportion matrix, which is defined as $\propmat_t=\adj_t \odot \textbf{Att}$.

After getting the activated proportion matrix, we can  calculate the latent traffic volume for all the road segments $\widehat{\obtallz}$ at time $t+1$ as:
\begin{equation}
\small
\setlength{\abovedisplayskip}{3pt}
\label{eq:pred}
    \widehat{\obtallz}_{t+1} = \propmat_t^\intercal \dot{\obtall}_{t} = (\adj_t \odot \textbf{Att})^\intercal \dot{\obtall}_{t}
\setlength{\belowdisplayskip}{3pt}
\end{equation}

\begin{figure}[t!]
\centering
  \begin{tabular}{c}
  \includegraphics[width=.89\linewidth]{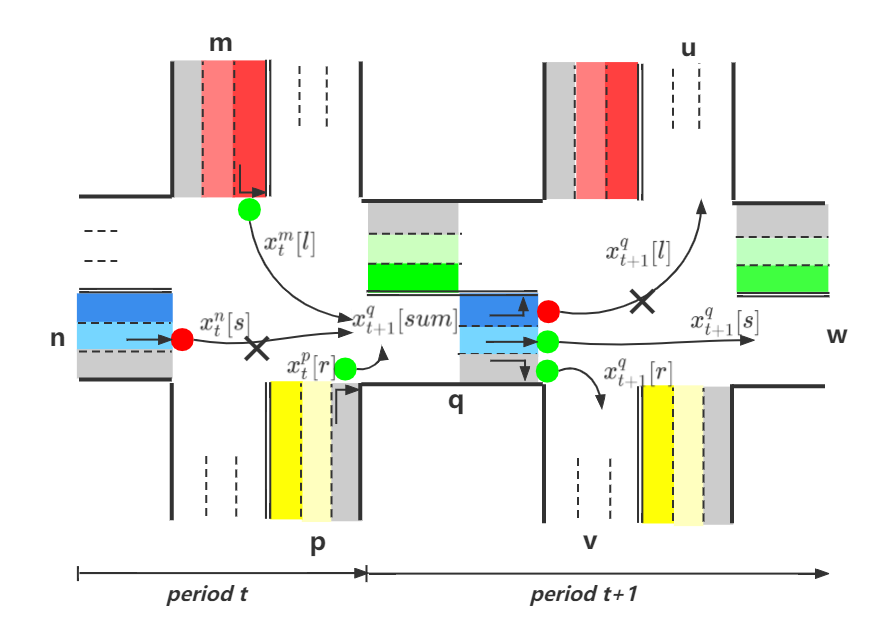}
  \end{tabular}
      \vspace{-3mm}
 \caption{The transition of traffic flows for road $q$.}
    \label{fig:transition}
    \vspace{-5mm}
\end{figure}

As we will show in the next section, Eq.~\eqref{eq:pred} is actually supported by the fundamental transition equations from transportation. 

\vspace{-4mm}
~\noindent\\~\textbf{Transition equations from transportation.}
From the perspective of transportation~\cite{aboudolas2009store,varaiya2013max}, the traffic flow on a road segment consists of incoming and outgoing flow. In Figure~\ref{fig:transition}, the traffic flows on road segment $q$ can be formulated as:
\begin{equation}
\small
\setlength{\abovedisplayskip}{3pt}
\label{eq:transition-eq-all}
    \mathbf{x}_{t+1}^{q}[sum] = \mathbf{x}_{t}^{q}[sum] - \mathbf{x}_{t}^{q}[out] + \mathbf{x}_{t+1}^{q}[in]   \\
\setlength{\belowdisplayskip}{3pt}
\end{equation}
where $\mathbf{x}_{t+1}^{q}[sum]$ denotes the total volume of traffic flow on road segment $q$ at time $t+1$, $\mathbf{x}_{t}^{q}[out]$ denotes the traffic volume leaving road segment $q$ at time $t$, and $\mathbf{x}_{t+1}^{q}[in]$ denotes the traffic volume that will arrive $q$ at time $t+1$.

Since the traffic flow is directional with downstream traffic flow coming from its upstream, which can be formulated as: 
\begin{equation}
\small
\setlength{\abovedisplayskip}{5pt}
\label{eq:transition-eq-all-1}
    \mathbf{x}_{t}^{q}[out] = \adj_t^{q,u} \cdot \gamma^{q,u} \cdot \mathbf{x}_{t}^{q}[l] + \adj_t^{q,w} \cdot \gamma^{q,w} \cdot \mathbf{x}_{t}^{q}[s] + \adj_t^{q,v} \cdot \gamma^{q,v} \cdot \mathbf{x}_{t}^{q}[r]   \\
\end{equation}
\begin{equation}
\small
\label{eq:transition-eq-all-2}
    \mathbf{x}_{t+1}^{q}[in] = \adj_t^{m,q} \cdot \gamma^{m,q} \cdot \mathbf{x}_{t}^{m}[l] + \adj_t^{p,q} \cdot \gamma^{p,q} \cdot \mathbf{x}_{t}^{p}[r] + \adj_t^{n,q} \cdot \gamma^{n,q} \cdot\mathbf{x}_{t}^{n}[s]  \\
\setlength{\belowdisplayskip}{3pt}
\end{equation}
where $u$, $v$, $w$ denote the downstream road segments of $q$, $\adj_t^{i,j}$ denotes the connectivity of road segment $i$ and $j$ based on the Phase-activated Adjacency Matrix at time $t$, $\adj_t^{i,j} = 1$ if road segment $i$ and $j$ is connected. $\gamma^{i,j}$ is the saturation flow rate in transportation~\cite{varaiya2013max,hurdle1984signalized}, which is a physical constant for $i$ and $j$, indicating the proportion of traffic volume that flows from $i$ to downstream $j$.

Intuitively, Eq.~\eqref{eq:transition-eq-all-1} indicates the outgoing flows are directional, moving from $q$ to its downstream roads that are permitted by traffic signals. Similarly, Eq.~\eqref{eq:transition-eq-all-2} indicates the incoming flows are also directional, moving from $q$'s upstream roads to $q$ are permitted by traffic signals. Specifically, as the proportion of traffic volume from $i$ to $j$, $\gamma^{i,j}$ considers the real-world situations where not all the intended traffic from $i$ to $j$ could be satisfied, due to physical road designs like speed limit, number of lanes, \etc~\cite{hurdle1984signalized}.

\vspace{-2mm}
\begin{theorem}[Connection with transition equations]
\label{theo}
The latent traffic volume calculated by Eq.~\eqref{eq:pred} equals to the transition equations, Eq.~\eqref{eq:transition-eq-all-1} and Eq.~\eqref{eq:transition-eq-all-2} from transportation.
\end{theorem}

\vspace{-3mm} 
\begin{proof}\renewcommand{\qedsymbol}{}
The proof can be done by expanding out the calculation for every element in $\obtallz_t$.

\vspace{-2mm} 
For any $\tau\in\{1,\dots,T\}$, the latent traffic state $\widehat{\obtallz}_{t-\tau+1}^i$ on road segment $i$ at time $t-\tau+1$ can be inferred by aggregating the features of all upstream traffic flows (allowed by traffic signal phase), which is similar to transition equations, Eq.~\eqref{eq:transition-eq-all-1} and Eq.~\eqref{eq:transition-eq-all-2}.
\begin{equation}
\small
\begin{aligned}
    \widehat{\obtallz}_{t-\tau+1}^i &=\sum_{j \in \mathcal{N}_{t-\tau}(i)}\adj_{t-\tau}^{j,i}\cdot \gamma^{j,i} \cdot \obtall_{t-\tau}^j \\
    &=\sum_{j \in \mathcal{N}_{t-\tau}(i)} (\adj_{t-\tau} \odot \gamma)^{j,i} \cdot \obtall_{t-\tau}^j 
\end{aligned}
\end{equation}
where $\mathcal{N}_t(i)$ denotes upstreams of road segment $i$. When we set $\gamma^{j,i} \in \textbf{Att}^\intercal$, the above equation equals to Eq.~\eqref{eq:pred}. Similar deduction also applys for Eq.~\eqref{eq:transition-eq-all-2}.

\end{proof}
\vspace{-6mm}

The transition equations ~\eqref{eq:transition-eq-all-1} and \eqref{eq:transition-eq-all-2} could be considered as the propagation and aggregation process of $\obtall_t$ toward $\widehat{\obtallz}_{t+1}$ under the guidance of Phase-activated Adjacency Matrix $\adj_t$, and the attention matrix $\textbf{Att}$ modeled by the base GNN module. According to Eq.~\eqref{eq:transition-eq-all},  a combination between $\obtall_{t}$ and $\widehat{\obtallz}_{t+1}$ could be applied for the final prediction for $\obtall_{t+1}$ to combine Eq.~\eqref{eq:transition-eq-all-1} and Eq.~\eqref{eq:transition-eq-all-2}.

\vspace{-1mm}
\subsubsection{Output Layer}
After \nt, an output layer is designed to model the combinations indicated by Eq.~\eqref{eq:transition-eq-all}. The output layer contains residual connections to comply with Eq.~\eqref{eq:transition-eq-all} and alleviate the over-smoothing problem. A fully connected layer is also applied to make sure the output of the network and the forecasting target have the same dimension. The final output is:
\begin{equation}
\small
\setlength{\abovedisplayskip}{3pt}
\label{eq:final}
    \widehat{\obtall}_{t+1} = ReLU( \sigma(\widehat{\obtallz}_{t+1} + \obtall_{t})\textbf{W}_L + \textbf{b}_L)
\setlength{\abovedisplayskip}{3pt}
\end{equation}
where $\textbf{W}_L$ and $\textbf{b}_L$ are learnable weights.

\vspace{-1mm}
\subsection{Iterative Imputation for Prediction}

Our model takes $T$ slices of $\obtall$ as input, and output two types of values: the prediction traffic volume of road segments in the historical $T-1$ time steps $(\widehat{\obtall}_{t-T+2},\widehat{\obtall}_{t-T+3},\cdots,\widehat{\obtall}_t)$ and states of all road segments in the next time step $\widehat{\obtall}_{t+1}$. 
At iteration $\tau$, the model would provide the predicted traffic volumes $\widehat{\obtall}_{t-\tau}$ on all the road segments for time $t-\tau+1$ , no matter accurate or not. Then the unobserved part from predicted traffic volume $\widehat{\ddot{\obtall}}_{t-\tau+1}$ will be merged with the observed states $\dot{\obtall}_{t-\tau+1}$ as merged states ${\obtall}'_{t-\tau+1}$ for $t-\tau+1$:
\begin{align}
\footnotesize
\label{eq:merge}
{\obtall}'_{t-\tau+1}=\dot{\obtall}_{t-\tau+1} + \widehat{\ddot{\obtall}}_{t-\tau+1}={\obtall}_{t-\tau+1}\odot\mask + \widehat{\obtall}_{t-\tau+1}\odot (1-\mask)
\end{align}
Then ${\obtall}'_{t-\tau+1}$, along with the imputed states from previous iterations $\{{\obtall}'_{t-T+1},\cdots, {\obtall}'_{t-\tau}\}$ and observed states for future iterations $\{\dot{\obtall}_{t-\tau+2},\cdots ,\dot{\obtall}_{t}\}$,  would be used to predict ${\widehat{\obtall}}_{t-\tau+2}$ for the next iteration. 
Since the first iteration do not have predictions from the past, random values are generated as $\widehat{\obtall}_{t-T+1}$. 
At the last iteration, we will have all the past merged states (${\obtall}'_{t-T+1}, {\obtall}'_{t-\tau+1},\cdots ,{\obtall}'_{t}$),
and the predicted value $\widehat{\obtall}_{t+1}$ will be used as the final prediction. 
~\noindent\\~\textbf{Loss function.}
Our loss function is proposed as follows:
\begin{align}
\small
\begin{split}\label{eq:loss}
    \min\limits_{\theta} \mathcal{L}_p(\theta) = & \frac{1}{T-1} \frac{1}{N}\sum_{i=1}^{T-1}\sum_{j=1}^{N}  \parallel(\textbf{x}^j_{T-i}-\widehat{\textbf{x}}^j_{T-i})\odot \mask^j\parallel_2 \\[-1ex]
    & + \frac{1}{N}\sum_{j=1}^{N} \parallel(\textbf{x}^j_{T+1}-\widehat{\textbf{x}}^j_{T+1})\odot \mask^j\parallel_2  \\
\end{split}
\end{align}
where $\theta$ are the model parameters to update, $T$ is the historical time step, and N is the number of all the road segments in the network. The first part of Eq.~\eqref{eq:loss} aims to minimize loss for imputation, and the second part aims to minimize loss for prediction. With the mask matrix $\mask$, our model is optimized over the observed tensor $\dot{\obtall}_j$, but with iterative imputation, the prediction on unobserved states could also be optimized. The detailed algorithm can be found in Appendix~\ref{app:algo}.

\section{Experiments}
In this section, we conduct experiments to answer the following research questions\footnote{The source codes are publicly available from ~\url{https://github.com/ShawLen/DTIGNN}.}: 

\noindent$\bullet$
\textbf{RQ1}: Compared with state-of-the-arts, how does \ours perform?

\noindent $\bullet$
\textbf{RQ2}: How do different components affect \ours?

\noindent $\bullet$
\textbf{RQ3}: Is \ours flexible to be integrated into existing methods?

\noindent $\bullet$
\textbf{RQ4}: How does \ours support downstream tasks?

\begin{table*}[t!]
\small
\centering
\caption{The MAE, RMSE and MAPE of different methods on synthetic data and real-world data. The lower, the better.}
\begin{tabular}{cccccccccc}
\toprule
Datasets              & Metrics  & STGCN~\cite{yu2017spatio}  & STSGCN~\cite{song2020spatial} & ASTGCN\cite{guo2019attention} & ASTGNN\cite{guo2021learning} & WaveNet~\cite{wu2019graph} & \begin{tabular}[c]{@{}c@{}}Ours \\ (ASTGNN)\end{tabular}  & \begin{tabular}[c]{@{}c@{}}Ours \\ (WaveNet)\end{tabular}  \\ \midrule
\multirow{3}{*}{\DUni} & MAE  & 0.0563 & 0.1244 & 0.0605 & 0.0562 & 0.0427       & 0.0484        & \textbf{0.0378}              \\
                      & RMSE & 0.1885 & 0.2993 & 0.2092 & 0.2124 & 0.1981       & 0.1920        & \textbf{0.1825}              \\
                      & MAPE & 0.0216 & 0.0460 & 0.0302 & 0.0256 & 0.0165       & 0.0215        & \textbf{0.0145}              \\ \midrule
\multirow{3}{*}{\DHZ}  & MAE  & 0.4909 & 0.6079 & 0.4458 & 0.4020 & 0.4556       & \textbf{0.3810}        & 0.4071              \\
                      & RMSE  & 0.8756 & 0.9104 & 0.7425 & 0.7408 & 0.8668       & \textbf{0.6618}        & 0.6883              \\
                      & MAPE  & 0.3135 & 0.3863 & 0.2953 & 0.2527 & 0.2987       & \textbf{0.2455}        & 0.2599              \\ \midrule
\multirow{3}{*}{\DNY}  & MAE  & 0.2651 & 0.4476 & 0.3136 & 0.2437 & \textbf{0.2168}       & 0.2437        &  0.2306                   \\ 
                      & RMSE  & 1.1544 & 1.1235 & 1.0625 & 1.0704 & 1.1485       & \textbf{0.9493}        &   1.1002                  \\
                      & MAPE  & 0.1146 & 0.2358 & 0.1620 & 0.1272 & 0.0988       & 0.1283        &   \textbf{0.1207}                \\
\bottomrule
\end{tabular}
\vspace{-3mm}
\label{tab:overall}
\end{table*}

\vspace{-1mm}
\subsection{Datasets}
We verify the performance of \ours on a synthetic traffic dataset \DUni and two real-world datasets, \DHZ and \DNY, where the road-segment level traffic volume are recorded with traffic signal phases. We generate sparse data by randomly masking several intersections out to mimic the real-world situation where the roadside surveillance cameras are not installed at certain intersections. Detailed descriptions and statistics can be found in Appendix~\ref{app:dataset}.
~\noindent\\~$\bullet$~\DUni. The synthetic dataset is generated by the open-source microscopic traffic simulator CityFlow~\cite{zhang2019cityflow}. The traffic simulator takes $4 \times 4$ grid road network and the traffic demand data as input, and simulates vehicles movements and traffic signal phases. 
The dataset is collected by logging the traffic states and signal phases provided by the simulator at every 10 seconds and randomly masking out several intersections.
~\noindent\\~$\bullet$~\DHZ. This is a public traffic dataset that covers a $4 \times 4$ network in Hangzhou, collected from surveillance cameras near intersections in 2016. This dataset records the position and speed of every vehicle at every second. Then we aggregate the real-world observations into 10-second intervals on road-segment level traffic volume, which means there are 360 time steps in the traffic flow for one hour. We treat these records as groundtruth observations and sample observed traffic states by randomly masking out several intersections.
~\noindent\\~$\bullet$~ \DNY. This is a public traffic dataset that covers a $7\times 28$ network in New York, collected by both camera data and taxi trajectory data in 2016. We sampled the observations in a similar way as in \DHZ.

\vspace{-1mm}
\subsection{Baseline Methods}

We compare our model mainly on data-driven baseline methods. For fair comparison, unless specified, all the methods are using the same data input, including the observed traffic volumes and traffic signal information.
~\noindent\\~$\bullet$~Spatio-Temporal Graph Convolution Network (STGCN)~\cite{yu2017spatio}: STGCN is a data-driven model that utilizes graph convolution and 1D convolution to capture spatial and temporal dependencies respectively.
~\noindent\\~$\bullet$~Spatial-Temporal Synchronous Graph Convolutional Networks (STSGCN)~\cite{song2020spatial}: STSGCN utilizes multiple localized spatial-temporal subgraph modules to capture the localized spatial-temporal correlations directly.
~\noindent\\~$\bullet$~Attention based Spatial Temporal Graph Convolutional Networks (ASTGCN)~\cite{guo2019attention}: ASTGCN uses spatial and temporal attention mechanisms to model spatial-temporal dynamics.
~\noindent\\~$\bullet$~Attention based Spatial Temporal Graph Neural Networks (ASTGNN)~\cite{guo2021learning}: Based on ASTGCN, ASTGNN further uses a dynamic graph convolution module to capture dynamic spatial correlations.
~\noindent\\~$\bullet$~GraphWaveNet (WaveNet)~\cite{wu2019graph}: WaveNet combines adaptive graph convolution with dilated casual convolution to capture spatial-temporal dependencies.


\vspace{-1mm}
\subsection{Experimental Settings and Metrics}
Here we introduce some of the important experimental settings and detailed hyperparameter settings can be found in Appendix~\ref{app:hyper}. 
We split all datasets with a ratio 6: 2: 2 into training, validation, and test sets. 
We generate sequence samples by sliding a window of width $T+T^{\prime}$. Specifically, each sample sequence consists of 31 time steps, where 30 time steps are treated as the input while the last one is regarded as the groundtruth for prediction.

Following existing studies~\cite{guo2019attention,guo2021learning,li2017diffusion,song2020spatial,yao2019revisiting}, MAE, RMSE and MAPE are adopted to measure the errors between predicted against groundtruth traffic volume. Their detailed formula for calculation can be found in Appendix~\ref{app:metrics}.
\vspace{-1mm}
\subsection{Overall Performance (RQ1)}

In this section, we investigate the performance of \ours on predicting the next traffic states under both synthetic data and real-world data. Table~\ref{tab:overall} shows the performance of \ours, traditional transportation models and state-of-the-art data-driven methods. We have the following observations:
~\noindent\\~$\bullet$~Among the data-driven baseline models, WaveNet and ASTGNN perform relatively better than other baseline methods in \DUni and \DNY. One reason is that these two models have a special design to capture the dynamic spatial graph, which could explicitly model the dynamic connectivity between road segments. 
~\noindent\\~$\bullet$~\ours can achieve consistent performance improvements over state-of-the-art prediction methods, including the best baseline models like Wavenet and ASTGNN. This is because \ours not only models the dynamic connectivity explicitly with the dynamic adjacency matrix, but also incorporates transition equations and imputation in the training process, the influences of missing observations are further mitigated. As we will see in Section~\ref{sec:sensitvity}, our proposed method can be easily integrated into the existing baseline methods and achieve consistent improvements.

\vspace{-1mm}
\subsection{Model Analysis}
To get deeper insights into \ours, we also investigate the following aspects about our model with empirical results: (1) ablation study on different parts of proposed components, (2) the effectiveness of imputation with prediction in one model, and (3) the sensitivity of \ours on different base models and under different data sparsity.

\begin{figure}[t!]
\small
\centering
  \begin{tabular}{ccc}
  \includegraphics[width=.28\linewidth]{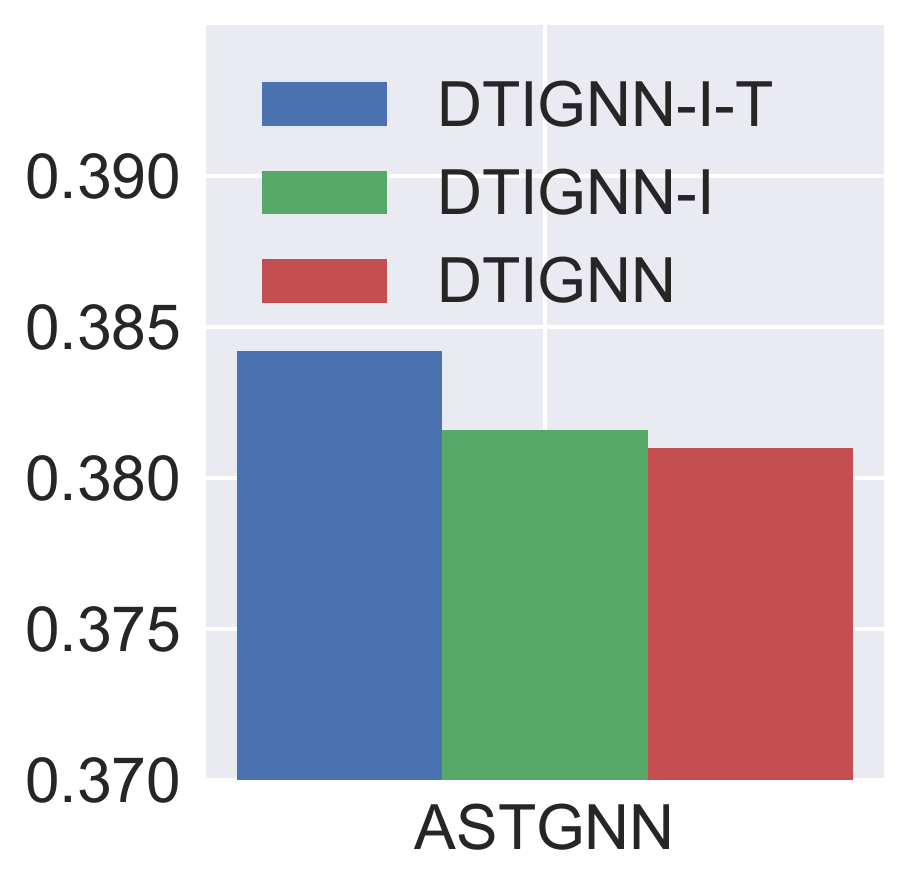} &
  \includegraphics[width=.28\linewidth]{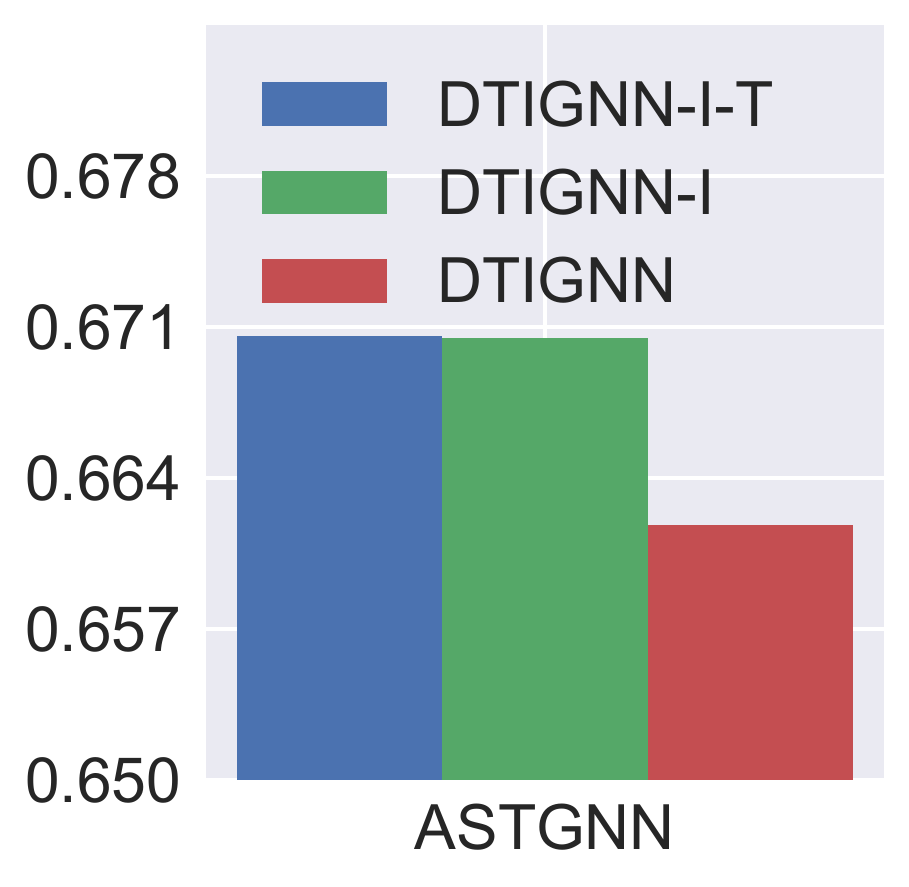} &
  \includegraphics[width=.28\linewidth]{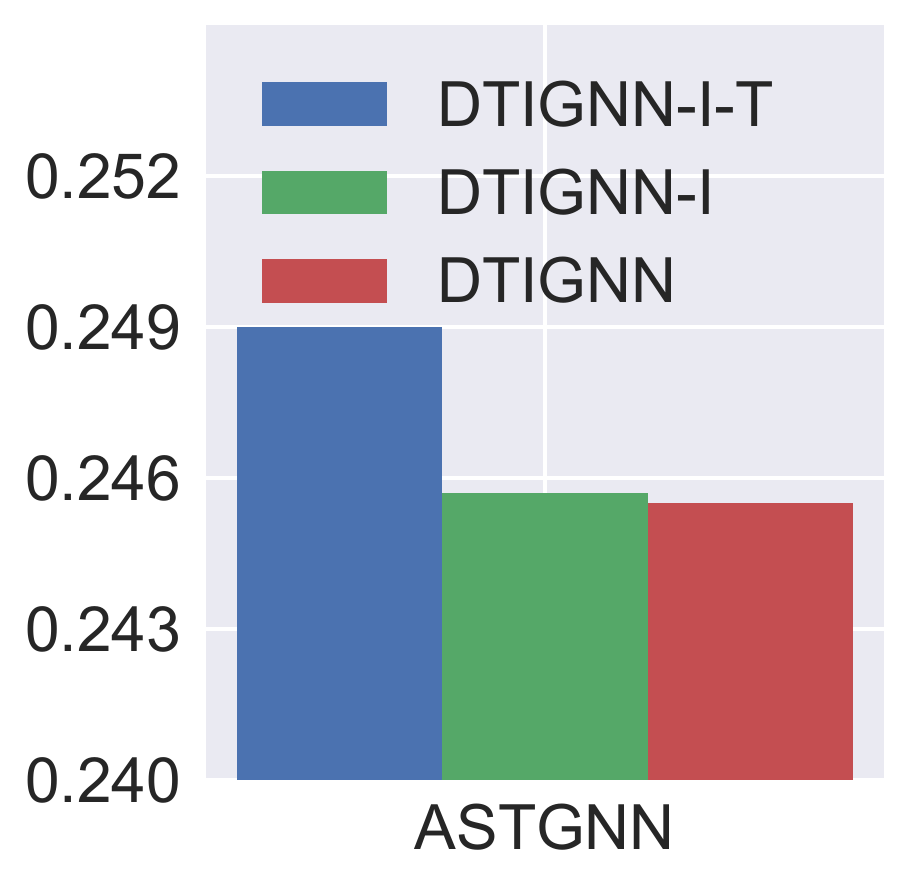}\\
  (a) MAE & (b) RMSE  & (c) MAPE  \\
  \end{tabular}
  \vspace{0.4mm}
 \caption{MAE, RSME and MAPE of \ours with the influence of different components on \DHZ. The lower, the better.}
    \label{fig:ablation}
    \vspace{-5mm}
\end{figure}

\begin{figure*}[tbh]
\centering
  \begin{tabular}{ccc}
  \includegraphics[width=.29\linewidth]{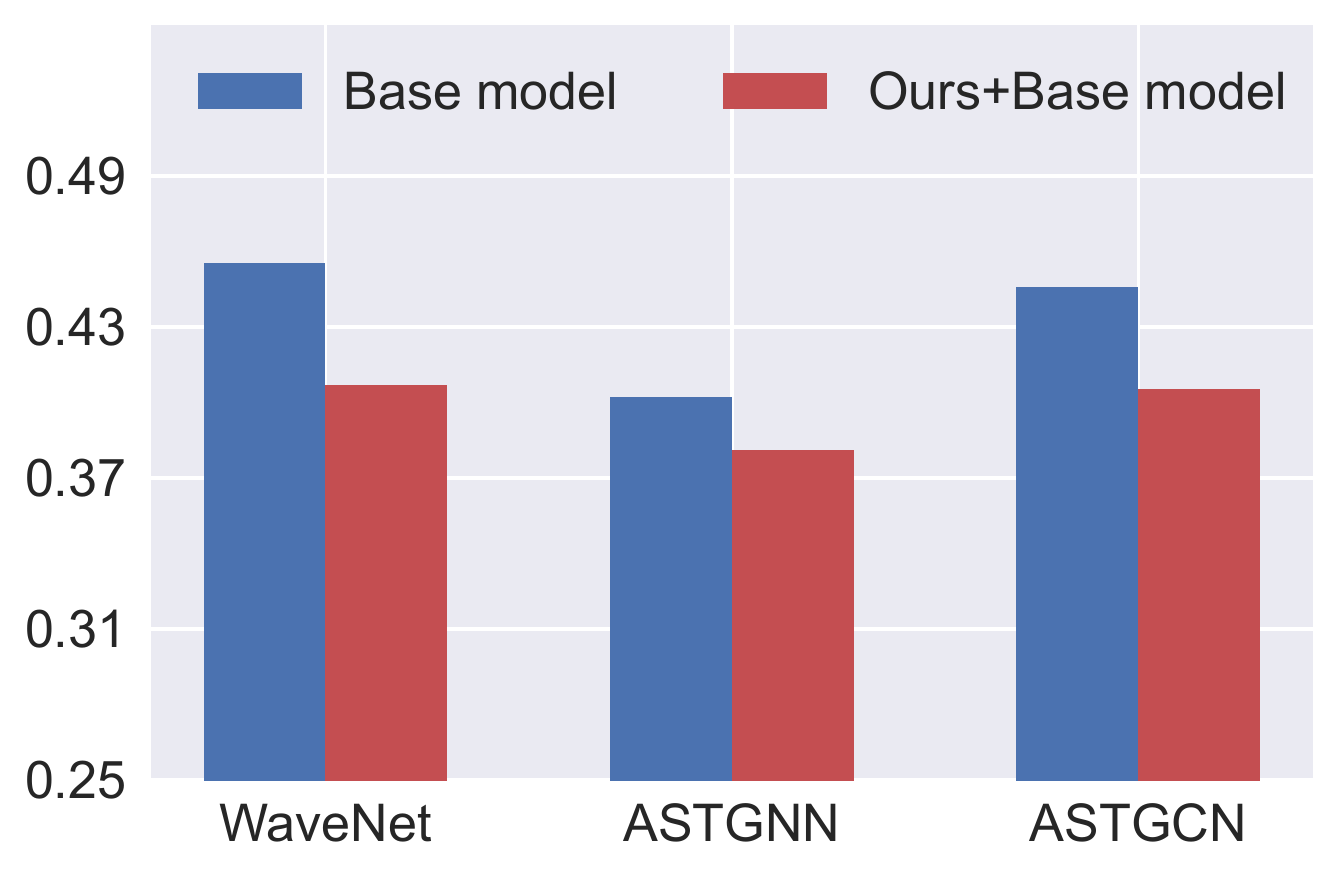} &
  \includegraphics[width=.29\linewidth]{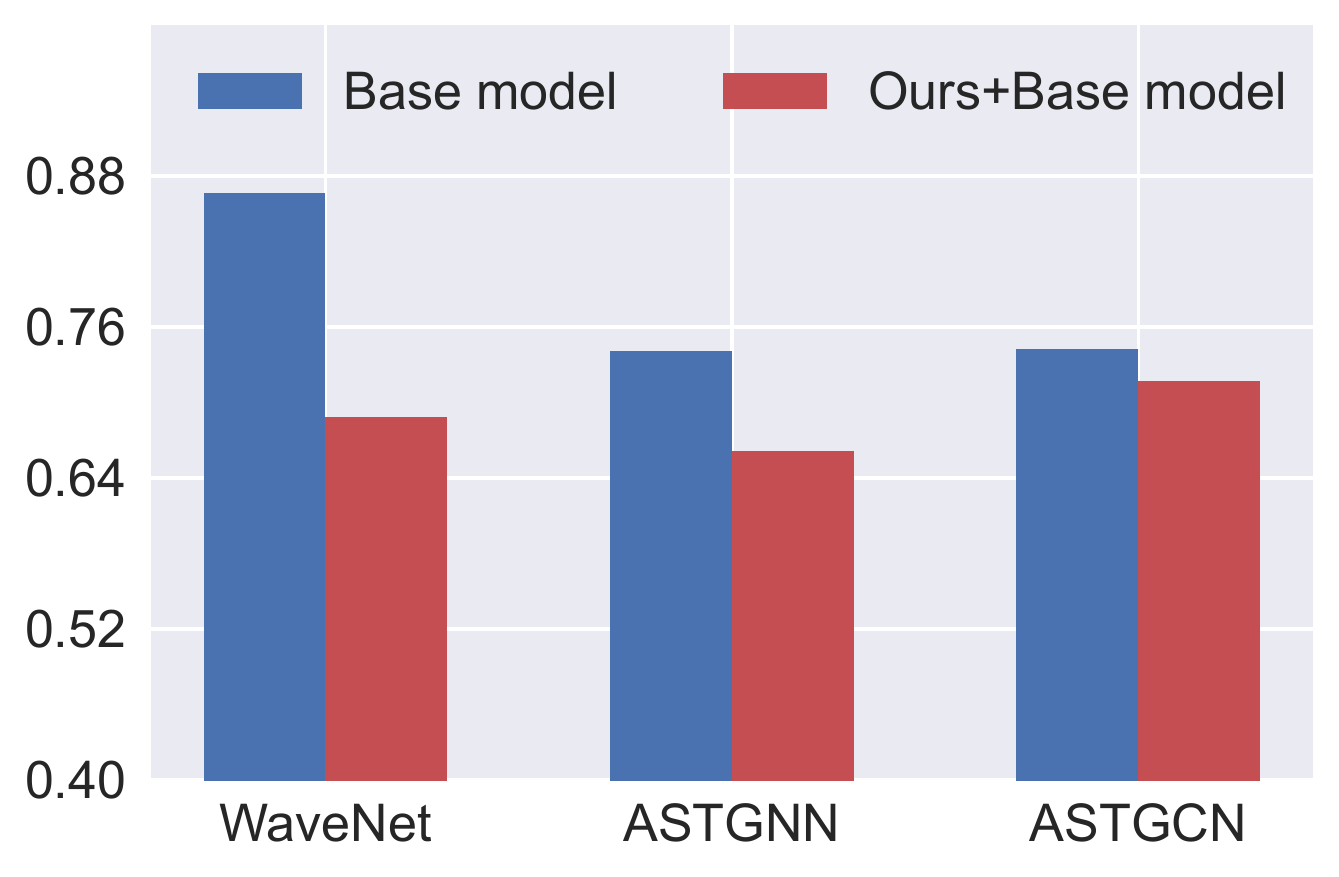} &
  \includegraphics[width=.29\linewidth]{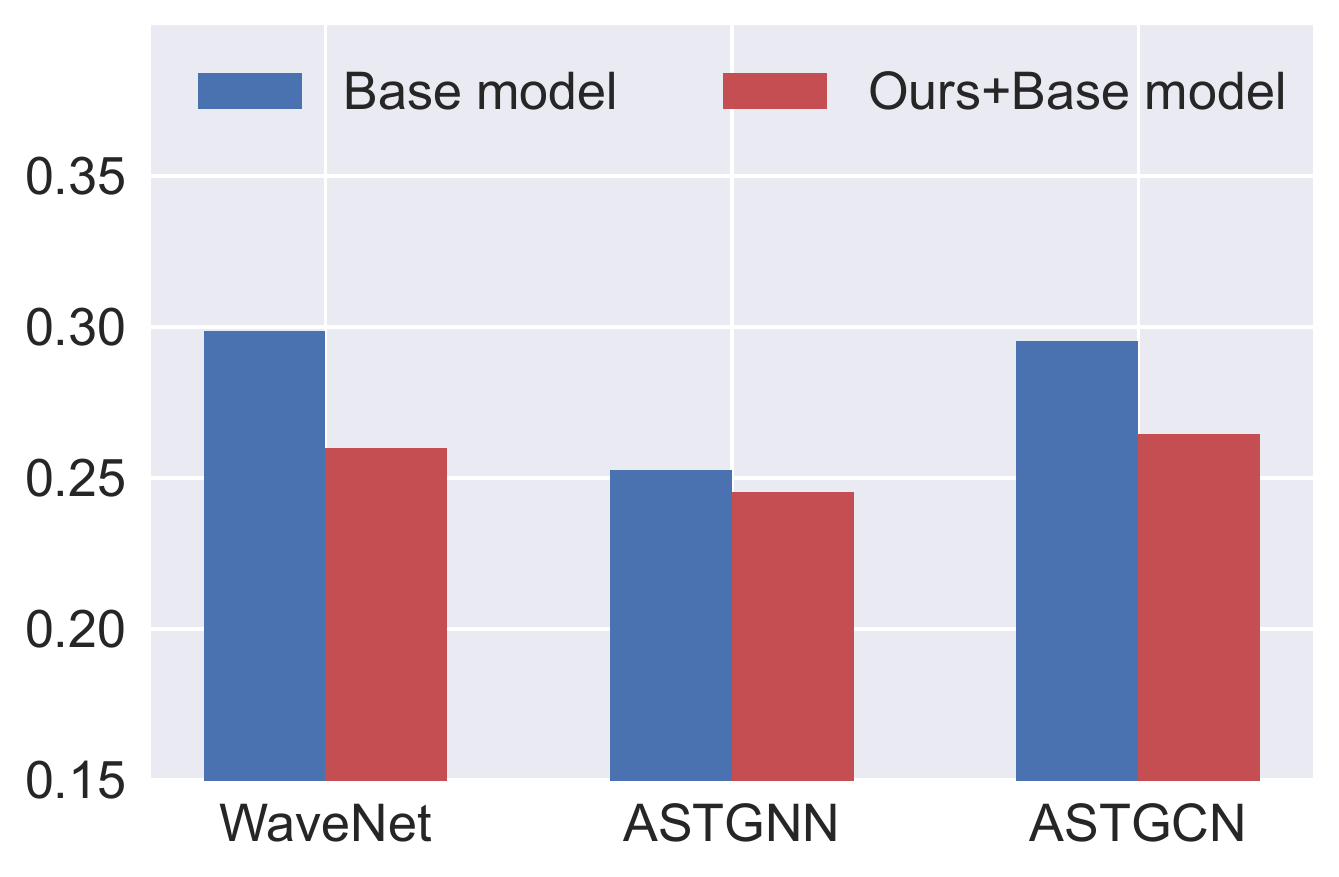}\\
  (a) MAE & (b) RMSE  & (c) MAPE  \\
  \end{tabular}
 \caption{MAE, RSME and MAPE of the variants of \ours against different baseline models on \DHZ. The lower, the better. The variants of \ours are implemented on corresponding baseline models. \ours achieves the better performance over corresponding baseline models. Similar trends are also found on other datasets.}
 \vspace{-3mm}
    \label{fig:multi-step}
\end{figure*}

\begin{figure*}[tbh]
\centering
  \begin{tabular}{ccc}
  \includegraphics[width=.29\linewidth]{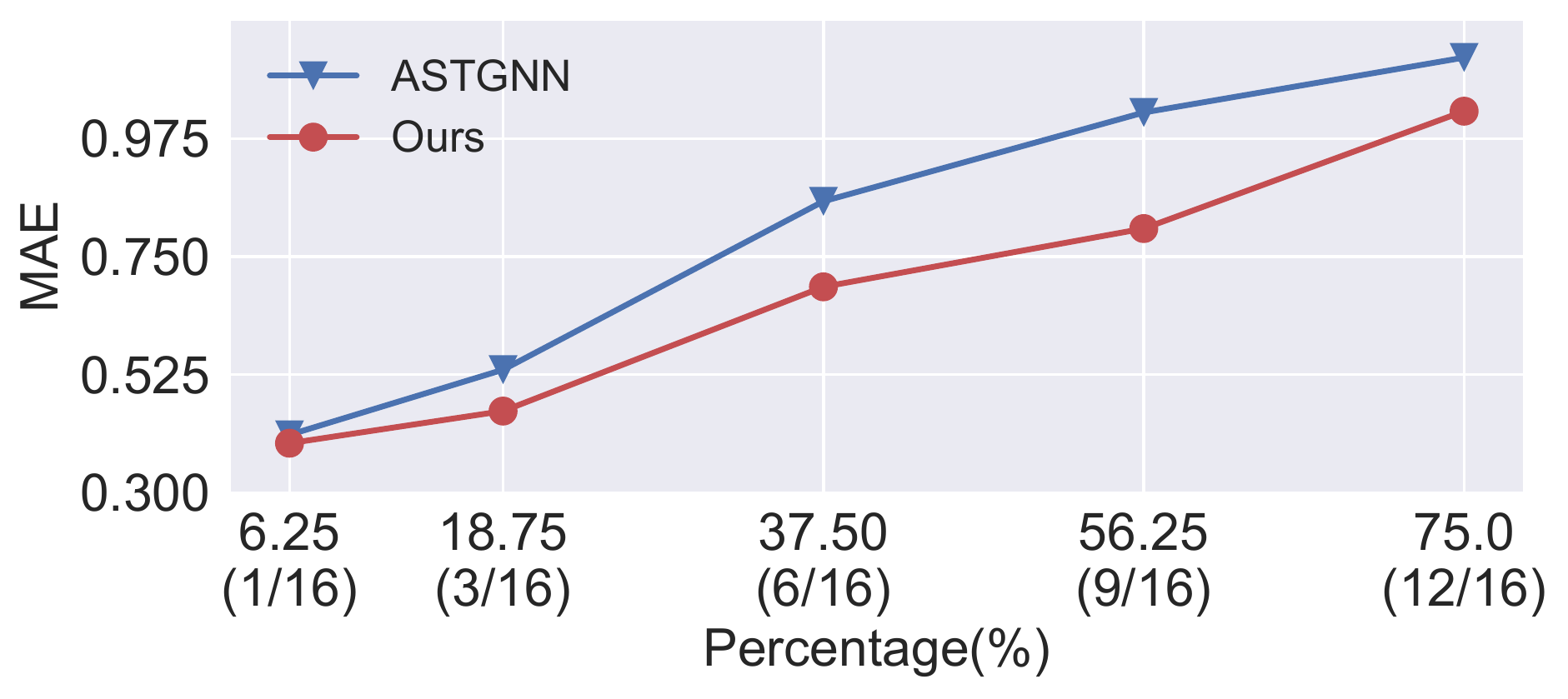} &
  \includegraphics[width=.28\linewidth]{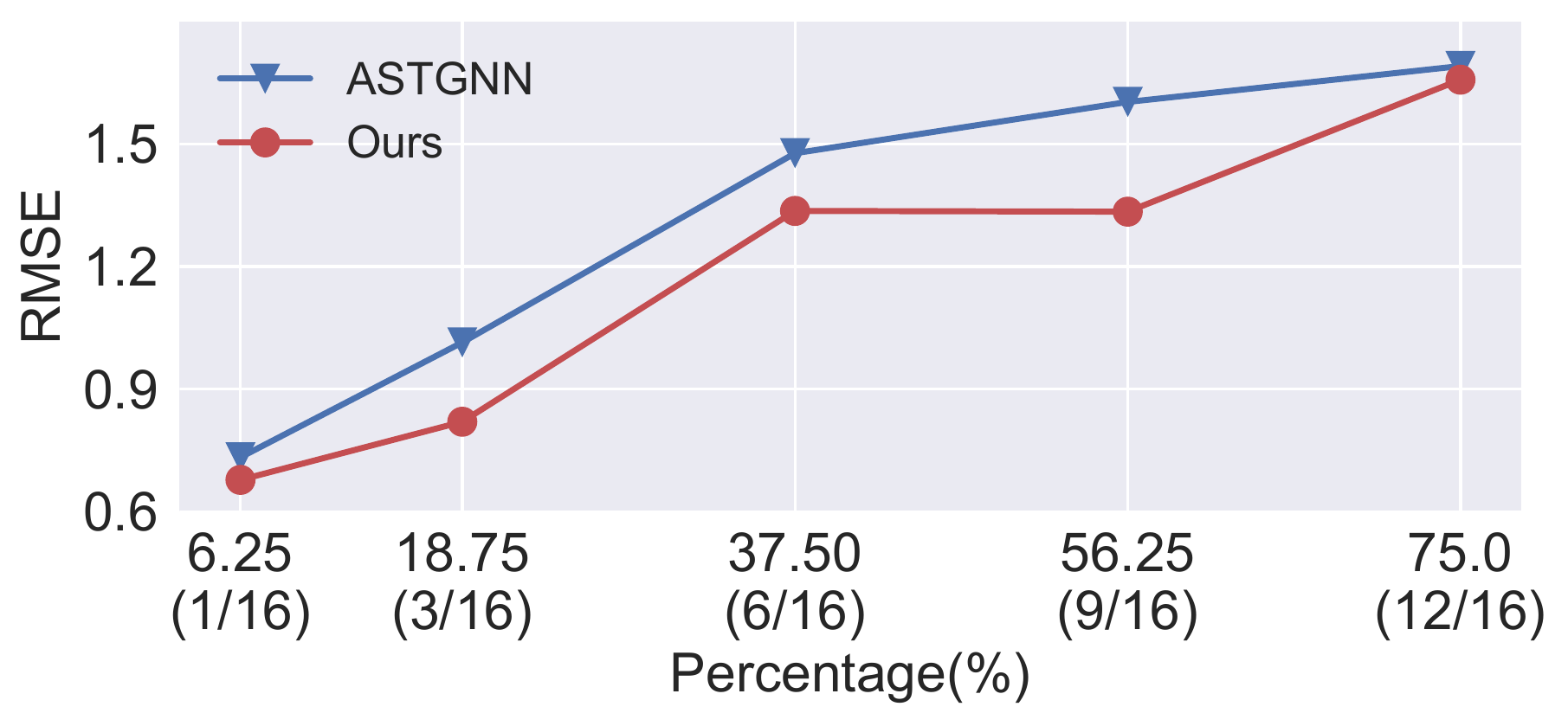} &
  \includegraphics[width=.29\linewidth]{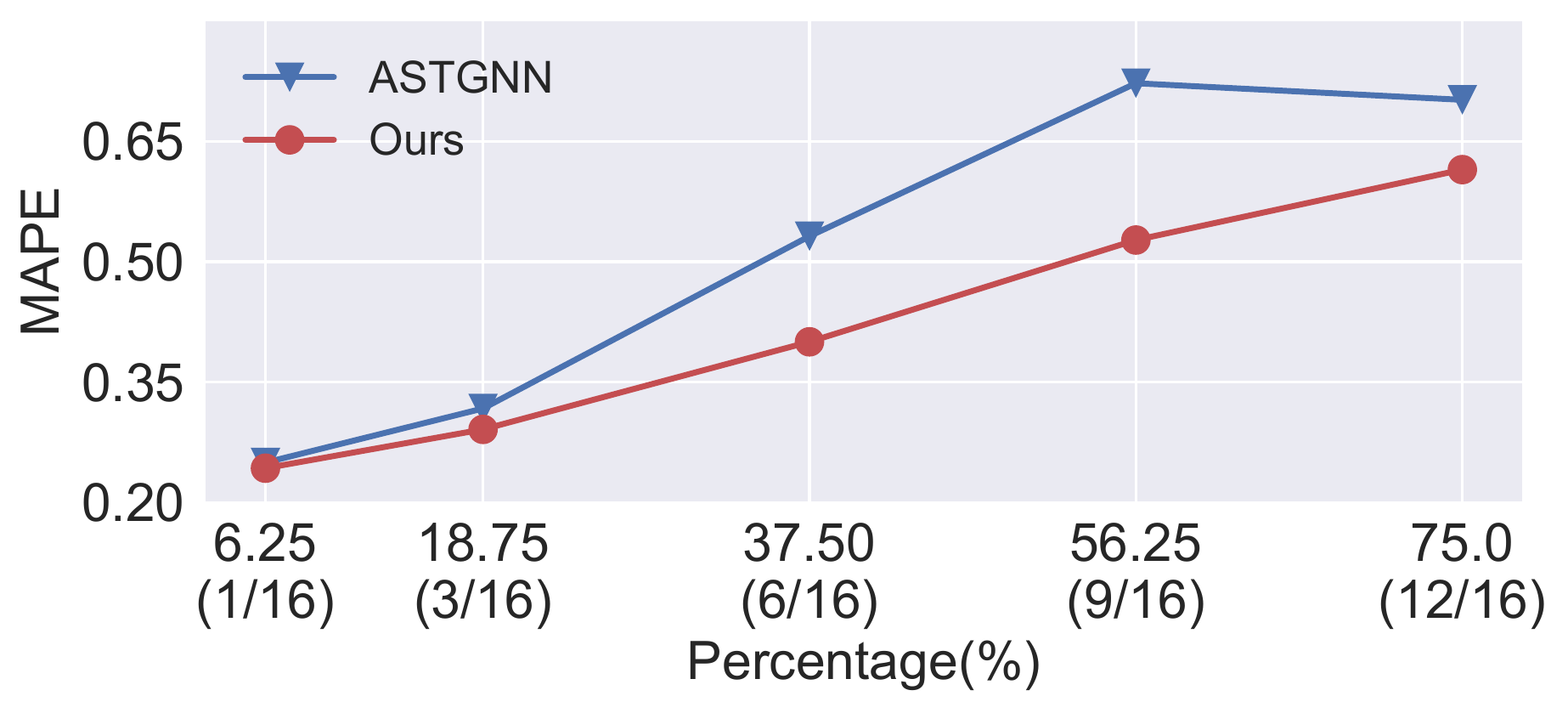}\\
  \vspace{-1mm}
  (a) MAE & (b) RMSE  & (c) MAPE  \\
  \end{tabular}
 \caption{MAE, RSME and MAPE of \ours against baseline models under different levels of sparsity on \DHZ. The lower, the better. The x-axis indicates the percentage of sparsity, which is calculated with the number of unobserved intersections divided by total number of intersections in the network (here \DHZ has 16 intersections) in percentage. Our proposed model achieves consistent better performance under different sparsity.}
    \label{fig:sparsity}
    \vspace{-5mm}
\end{figure*}

\vspace{-1mm}
\subsubsection{Ablation Study (RQ2)}
In this section, we perform an ablation study to measure the importance of dynamic graph and imputation on the following variants:
~\noindent\\~$\bullet$~\textbf{\oursNoPR}. This model takes both the network-level traffic volume and signal phases as its input and outputs the future traffic volume directly, which does not include \nt to model the transition equations and does not have imputation in the training process. It can also be seen as the baseline model ASTGNN with additional node features of traffic signal phases. 
~\noindent\\~$\bullet$~\textbf{\oursNoR}. Based on \oursNoPR, this model additionally uses \nt to incorporate transition equations, while it does not impute the sparse data in the training process. 
~\noindent\\~$\bullet$~\textbf{\ours}. This is our final model, which integrates \nt, and imputes sparse observations in the training process. 

Figure~\ref{fig:ablation} shows the performance of the variants. We can observe that \oursNoR outperforms \oursNoPR, indicating the effectiveness of \nt as a way to incorporate transition equations. \ours further outperforms \oursNoR, since it iteratively imputes the sparse data in the training process and optimize the imputation and prediction at the same time.

\begin{table}[h!]
\small
\centering
\caption{Performance of \ours against different baselines and their corresponding two-step variants on \DHZ. The lower, the better. Vanilla baseline methods and \ours learn from sparse observations, while the two-step variants impute sparse observations first and train on the imputed data. \ours
achieves the best performance in all cases. Similar trends are also found on other datasets.}
\begin{tabular}{cccc}
\toprule
Method & MSE    & RMSE   & MAPE   \\ \midrule
ASTGCN (vanilla)  & 0.4458 & 0.7425 & 0.2953 \\
ASTGCN (two step)  & 0.4423 & 0.7417 & 0.2928 \\ \midrule
WaveNet (vanilla) & 0.4556 & 0.8668 & 0.2987 \\
WaveNet (two step)  & 0.4104 & 0.6977 & 0.2594 \\ \midrule
ASTGNN  (vanilla) & 0.4020  & 0.7408 & 0.2527 \\
ASTGNN (two step)  & 0.3842 & 0.6706 & 0.2490  \\ \midrule
Ours   & 0.3810  & 0.6618 & 0.2455 \\
\bottomrule
\end{tabular}
\vspace{-4mm}
\label{tab:two-step}
\end{table}

\vspace{-1mm}
\subsubsection{Imputation Study}
To better understand how imputation helps prediction,
we compare two representative baselines with their two-step variants. Firstly, we use a pre-trained imputation model to impute the sparse data following the idea of~\cite{tang2019joint,yi2019citytraffic}. Then we train the baselines on the imputed data.

Table~\ref{tab:two-step} shows the performance of baseline methods in \DHZ. We find out that the vanilla baseline methods without any imputation show inferior performance than the two-step approaches, which indicates the importance of imputation in dealing with missing states. 
\ours further outperforms the two-step approaches. This is because \ours learns imputation and prediction in one step, combining the imputation and prediction iteratively across the whole framework. Similar results can also be found in\DUni and \DNY, and we omit these results due to page limits.

\vspace{-1mm}
\subsubsection{Sensitivity Study}
\label{sec:sensitvity}
In this section, we evaluate the sensitivity of \ours with different base models and different data sparsity.
~\noindent\\~$\bullet$~\textbf{Choice of base model. (RQ3)} \ours can be easily integrated with different GNN structures. To better understand how different base models influence the final performance,
we compare \ours with different GNN models as base GNN modules against corresponding GNN models. Figure~\ref{fig:multi-step}
summarizes the experimental results. Our proposed method performs steadily under different base models across various metrics, indicating the idea of our method is valid across different base models. In the rest of our experiments, we use ASTGNN as our base model and compare it with other methods.
~\noindent\\~$\bullet$~ \textbf{Data sparsity.}
In this section, we investigate how different sampling strategies influence \ours. We randomly sample 1, 3, 6, 9, 12 intersections out of 16 intersections from \DHZ and treat them unobserved to get different sampling rates. We set the road segments around the sampled intersections as the missing segments and evaluate the performance of our model in predicting the traffic volumes for all the road segments. As
is shown in Figure~\ref{fig:sparsity}, with sparser observations, \ours performs better than ASTGNN with lower errors, though the errors of both methods increase. Even when over half the observations are missing, our proposed method can still predict future traffic states with lower errors, indicating the consistency of \ours under a wide range of sparsity.
\raggedbottom

\vspace{-1mm}
\subsection{Case Study (RQ4)}
To test the feasibility of using our proposed model in the downstream tasks, we conduct experiments on a traffic signal control task, where only part of traffic states in the road network is observable.  Ideally, adaptive signal control models take the full states as input and provide actions to the environment, which then executes the traffic signal actions from the control models. However, in the sparse data setting, the signal models at unobserved intersections will fail to generate any adaptive actions since there is no state as input. With a well-trained state transition model, the full traffic states could be predicted for all the intersections, upon which the signal control models could generate actions. \emph{Intuitively, a good state transition model would provide the signal controller with more precise states, and enable better results in traffic signal control.} 

\begin{figure}[t!]
\small
\centering
  \begin{tabular}{cc}
  \includegraphics[width=.38\linewidth]{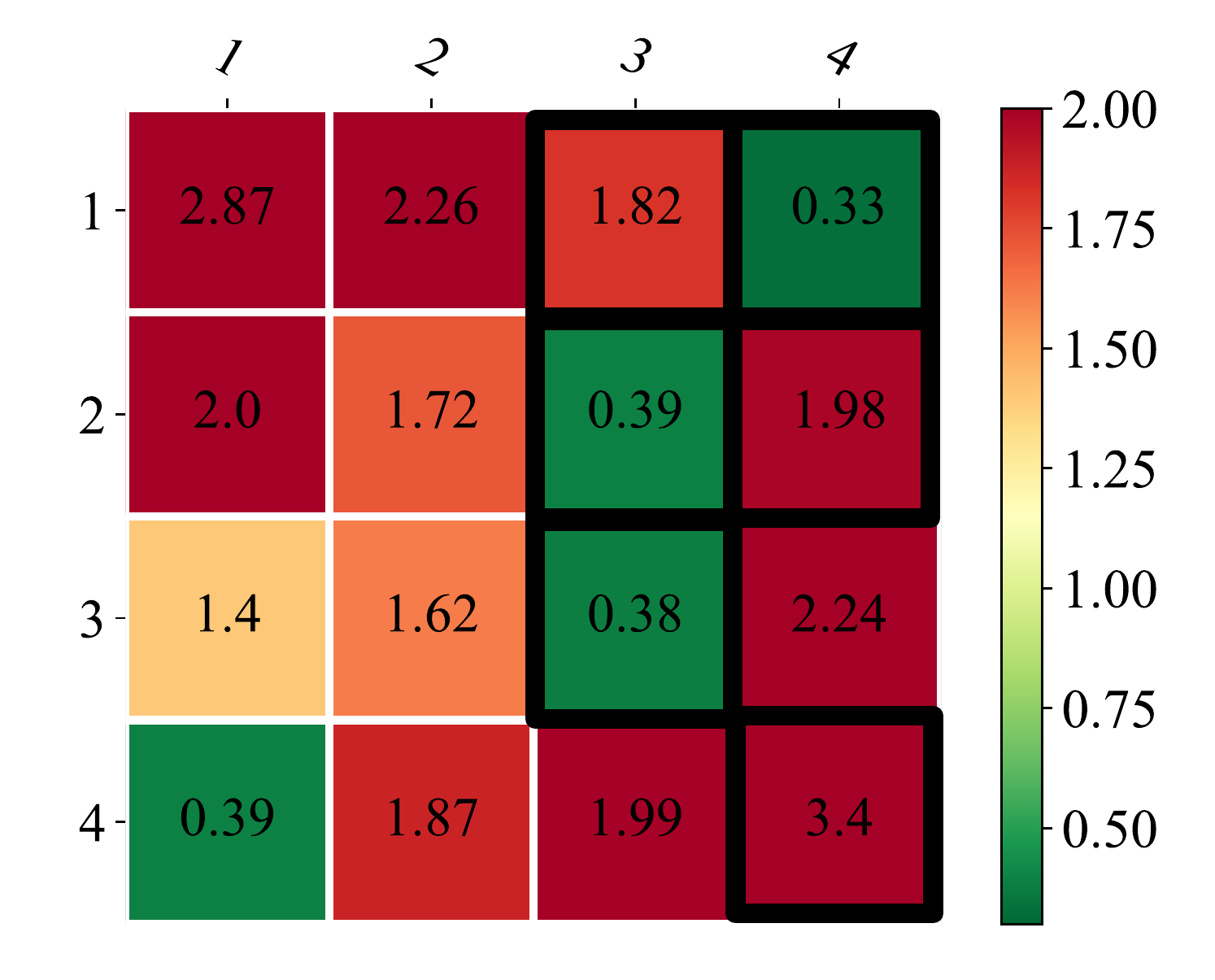} &
  \includegraphics[width=.38\linewidth]{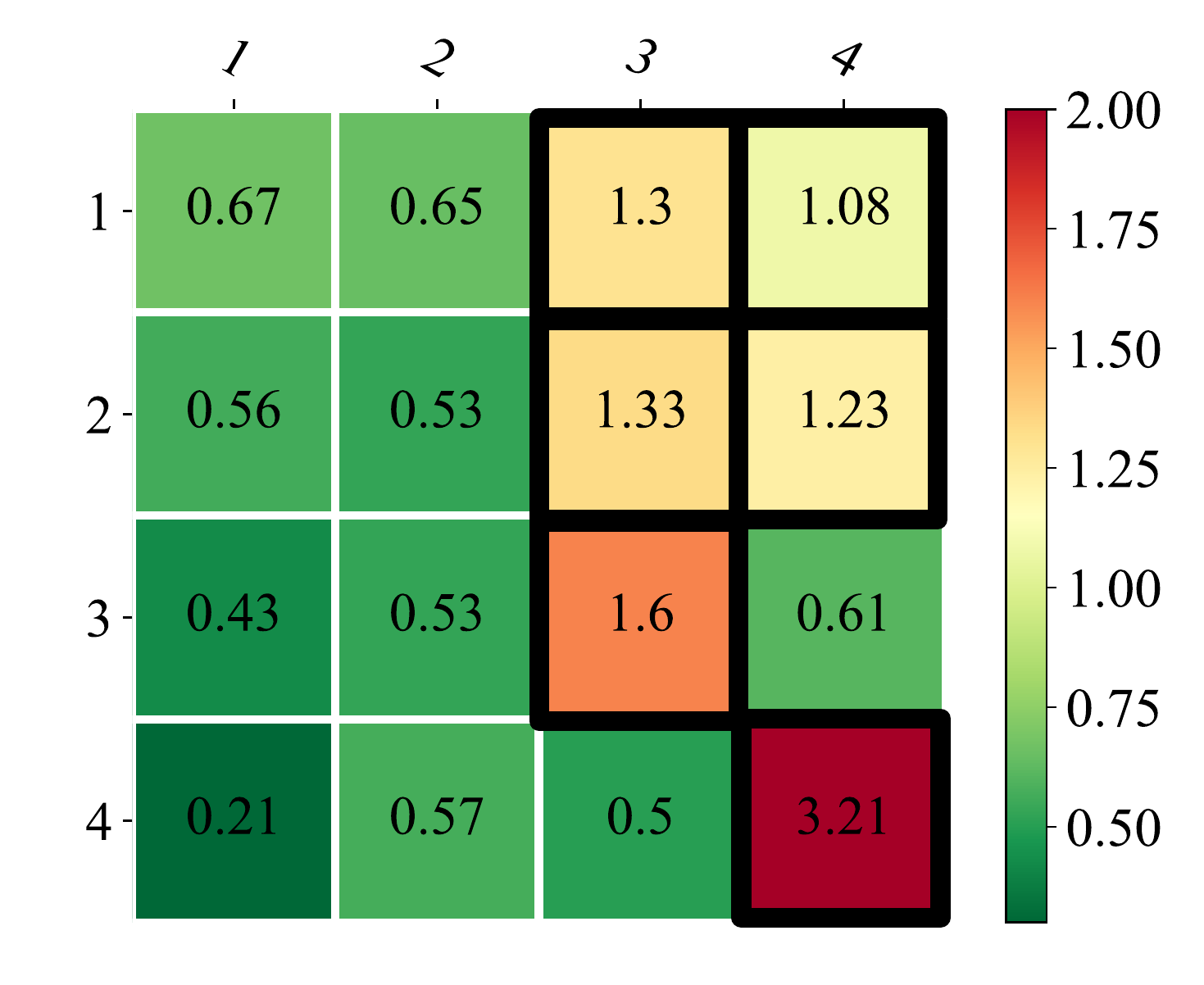}
   \\
  \multicolumn{2}{c}{(a) RMSE of baseline (left) and \ours (right). The lower, the better.} \\
 \includegraphics[width=.38\linewidth]{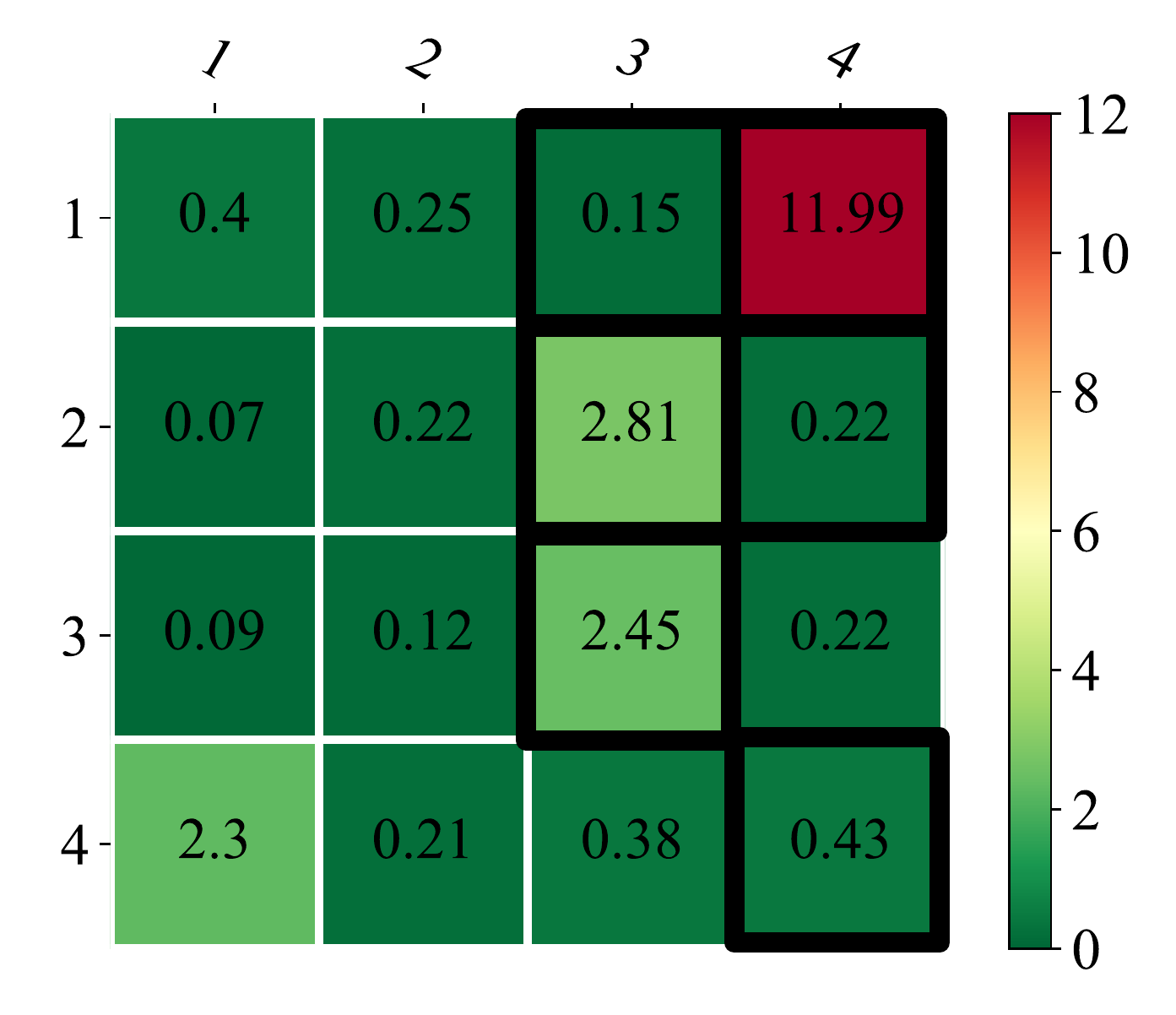} &
  \includegraphics[width=.38\linewidth]{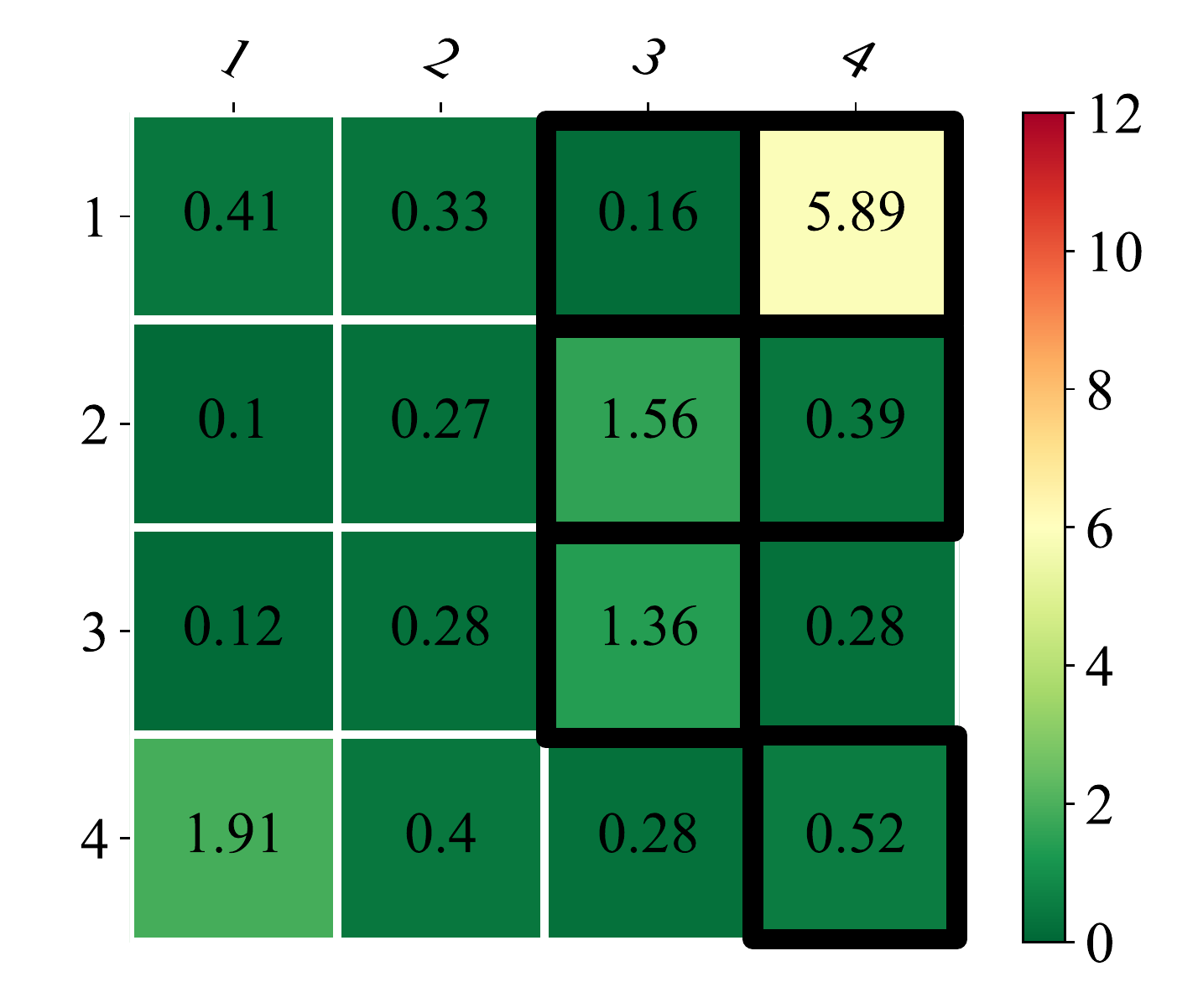} \\
  \multicolumn{2}{c}{\begin{tabular}[c]{@{}c@{}}(b) Queue length of MaxPressure using predictions from\\baseline (left) and \ours (right). The lower, the better. \end{tabular} }  \\
  \end{tabular}
 \caption{The performance of the baseline method and \ours under sparse data on \DHZ for two sequential tasks: traffic state modeling w.r.t. RMSE, and traffic signal control w.r.t. queue length using MaxPressure. The x-axis and y-axis indicate the location of the intersections in the road network of \DHZ. The black boxes indicate the intersections without any observation. \ours has lower prediction error in predicting next traffic states. These predictions are then used by MaxPressure~\cite{varaiya2013max} to achieve better performance with a lower average queue length. Best viewed in color.}
    \label{fig:case}
\vspace{-5mm}
\end{figure}

The experiment is conducted on the traffic simulator CityFlow~\cite{zhang2019cityflow} with \DHZ. We use the states returned by the traffic simulator as full states and mask out the observations of several intersections to mimic the sparse data setting. Based on the sparse data, state transition models can predict the full future states. Then the traffic signal control model (here we use an adaptive model, MaxPressure~\cite{varaiya2013max}) will decide the actions and feed the actions into the simulator. We compare the performance of \ours against ASTGNN, the best baseline model on \DHZ, for the state transition task and the traffic signal control task. For traffic signal control task, the average queue length of each intersection and the average travel time of all the vehicles are used as metrics, following the existing studies~\cite{wei2018intellilight,zhang2019cityflow}. Detailed descriptions can be found in Appendix~\ref{app:case}.

Figure~\ref{fig:case} shows the experimental results for each intersection in \DHZ, and we have the following observations:
~\noindent\\~$\bullet$~\ours has better performance in both predicting traffic states and traffic signal control in Figure~\ref{fig:case}(a) and Figure~\ref{fig:case}(b). We also measure the average travel time of all vehicles in the network and found out that the average travel time by \ours is 494.26 seconds, compared to the 670.45 seconds by ASTGNN. This indicates that \ours can better alleviate downstream signal control task that suffers from missing data problems.
~\noindent\\~$\bullet$~We also notice that, although ASTGNN shows lower RMSE on certain intersections (\eg, intersection 1\_4, 2\_3, 3\_3 on the left of Figure~\ref{fig:case}(a)), its performance on traffic signal control result is inferior. This is because network-level traffic signal control requires observation and coordination between neighboring intersections, which requires accurate modeling on global traffic states. In contrast, \ours shows satisfactory prediction performance on the global states and thus achieve better performance on traffic signal control task even when the traffic signal controller. 

\section{Conclusion}

In this paper, we propose a novel and flexible approach to model the network-level traffic flow transition.
Specifically, our model learns on the dynamic graph induced by traffic signals with a network design grounded by transition equations from transportation, and predicts future traffic states with imputation in the process. We conduct extensive experiments using synthetic and real-world data and demonstrate the superior performance of our proposed method over state-of-the-art methods. We further show in-depth case studies to understand how the state transition models help downstream tasks like traffic signal control, even for intersections without any traffic flow observations.

We would like to point out several important future directions to make the method more applicable to the real world. First, more traffic operations can be considered in the modeling of transitions, including road closures and reversible lanes. Second, the raw data for observation only include the phase and the number of vehicles on each lane. More exterior data like the road and weather conditions might help to boost model performance.


\bibliographystyle{ACM-Reference-Format}
\balance
\bibliography{sample-base}

\newpage

\appendix






\section{algorithm}
\label{app:algo}

\begin{algorithm}
\caption{Training procedure of DTIGNN}
\KwIn{Observed state tensor $\dot{\mathcal{X}}=(\dot{X}_{t-T+1}, \dot{X}_{t-T+2},\cdots, \dot{X}_{t})$, phase Activate Matrix $(\mathcal{A}_{t-T+1}, \mathcal{A}_{t-T+2},\cdots, \mathcal{A}_t)$, Static Adjacency Matrix $\mathcal{A}$, initial training options and model’s parameters $\theta_0$, \etc}
\KwOut{$X=(X_{t+1})$, the last dimension is 3 dimensions represent l,s,r traffic volumes of road segments,respectively.}
\For{$epoch\gets {0,1,\cdots}$}{
    \For{$\tau \gets {t-T+1,\cdots,t}$}{
    \tcp{Base GNN Module}
    Generate Attentation Matrix $Att$ using Eq~\eqref{eq:att}\;
    \tcp{Neural Transition Layer}
    Generate Proportion Matrix $\Gamma_t$\;
    Calculate latent states $\widehat{\obtallz}_{\tau}$ using Eq~\eqref{eq:pred}\;
    \tcp{Output Layer}
    Predict traffic states of unobserved road segments using Eq~\eqref{eq:final} \;
    Impute on $\ddot{\obtall}_{\tau}$ with Eq~\eqref{eq:merge}
    }
    Predict $X$\;
    Update $\theta$ by minimize the training loss with Eq~\eqref{eq:loss}
}
\end{algorithm}

\section{Dataset Description}
\label{app:dataset}

\begin{figure}[t!]
\centering
  \includegraphics[width=.9\linewidth]{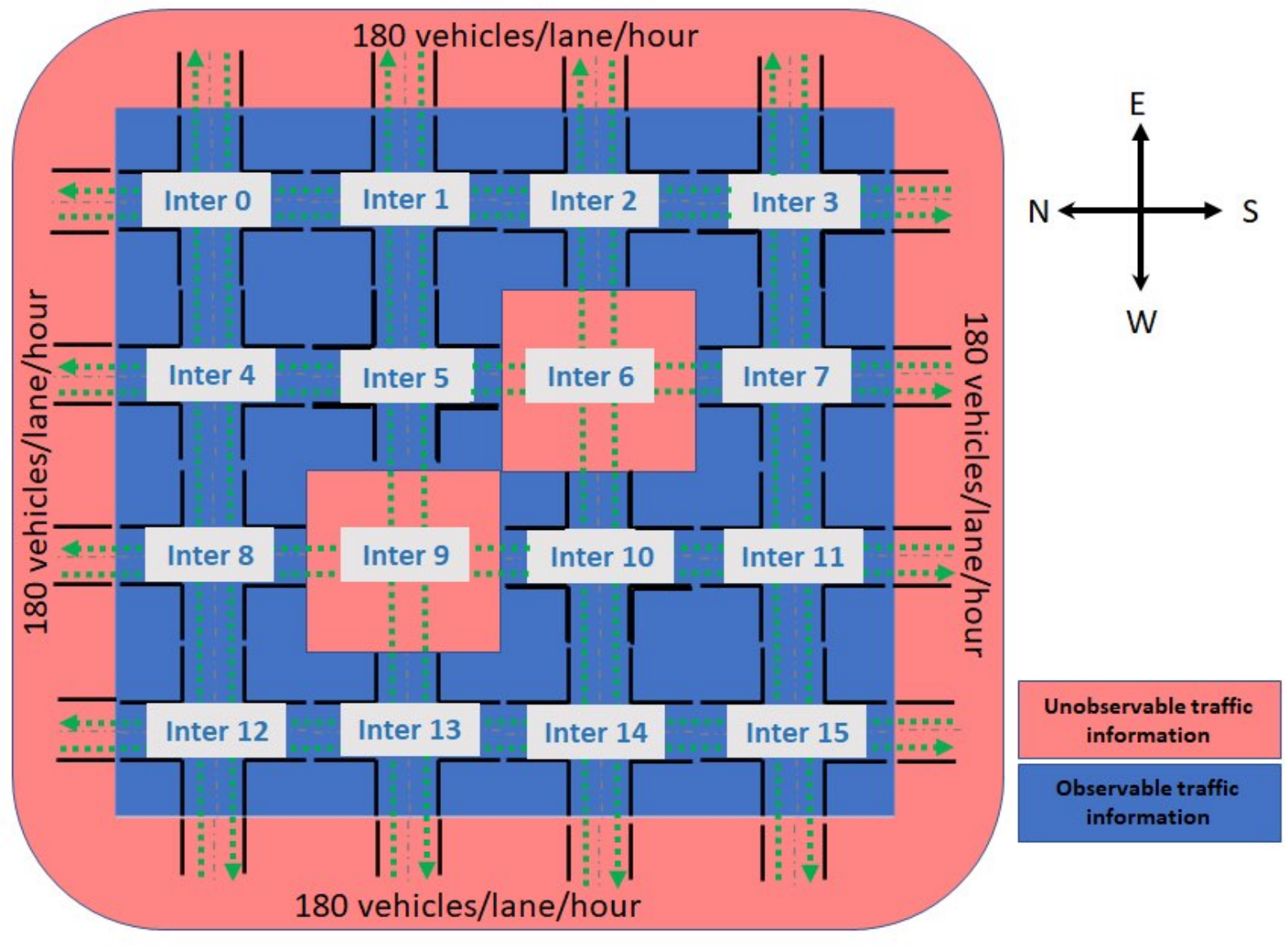}
 \caption{Road networks for synthetic data set. \DUni has 16 intersections with bi-directional traffic. Green dot lines represent traffic flow direction on each lane. Traffic information in the red regions is unobservable while in the  blue region information is entirely recorded.}
    \label{fig:Uniform}
    \vspace{-3mm}
\end{figure}

\begin{figure}[t!]
\centering
  \includegraphics[width=.95\linewidth]{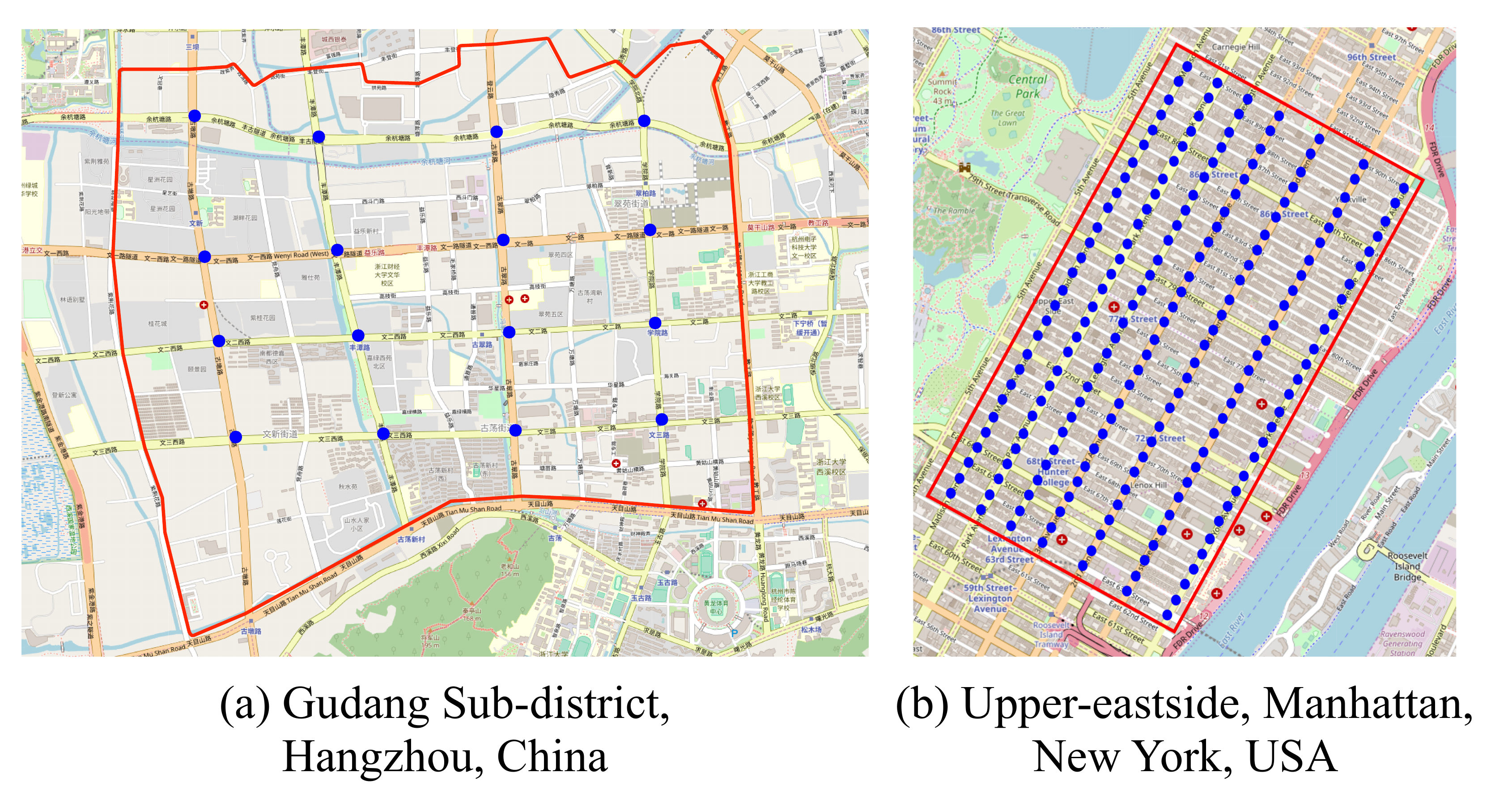}
 \caption{Road networks for real-world datasets. Red polygons are the areas we select to model, blue dots are intersections with traffic signals. (a) \DHZ, 16 intersections with uni- \& bi-directional traffic. (b) \DNY, 196 intersections with uni-directional traffic.}
    \label{fig:data}
    \vspace{-3mm}
\end{figure}

We conduct experiments on one synthetic traffic dataset \DUni and two real-world datasets, \DHZ and \DNY, where the road-segment level traffic volume are recorded with traffic signal phases. The data statistics are shown in Table~\ref{tab:data}, and the road networks in these datasets are shown in Figure~\ref{fig:Uniform} and Figure~\ref{fig:data}.

\begin{table}[htb]
\centering
\caption{Statistics of datasets.}
\label{tab:data}
\begin{tabular}{cccc}
\toprule
   Dataset    & \DUni & \DHZ & \DNY \\ \midrule
Duration (seconds)   &   3600  &  3600   &  3600  \\
Time steps   &   360  &  360   &  360  \\
\# of intersections   &   16  &  16   &  196 \\
\# of road segments   &   80  &  80   &  854 \\
\begin{tabular}[c]{@{}c@{}} \# of groundtruth states (full)\end{tabular}& 23040   & 23040 & 282240 \\
\begin{tabular}[c]{@{}c@{}} \% of unobserved intersections\end{tabular}& 12.5   & 12.5 & 10.4 \\
\bottomrule
\end{tabular}
\vspace{-5mm}
\end{table}

~\noindent\\~$\bullet$~\DUni. The synthetic dataset is generated by the open-source microscopic traffic simulator CityFlow~\cite{zhang2019cityflow}. The traffic simulator takes $4 \times 4$ grid road network and the traffic demand data (represented as the origin and destination location, the route and the start time) as input, and simulates vehicles movements and traffic signal phases. 
Each intersection in the road network has four directions (West$\rightarrow$East, East$\rightarrow$West, South$\rightarrow$North, and North$\rightarrow$South), and 3 lanes (300 meters in length and 3 meters in width) for each direction. Vehicles come uniformly with 180 vehicles/lane/hour in both West$\leftrightarrow$East direction and  South$\leftrightarrow$North direction. The dataset is collected by logging the traffic states and signal phases provided by the simulator at every 10 seconds and randomly masking out several intersections.
~\noindent\\~$\bullet$~\DHZ. This is a public traffic dataset that covers a $4 \times 4$ network of Gudang area in Hangzhou, collected from surveillance cameras near intersections in 2016. This dataset records the position and speed of every vehicle at every 1 second. Then we aggregate the real-world observations into 10-second intervals on road-segment level traffic volume, which means there are 360 time steps in the traffic flow for one hour. We treat these records as groundtruth observations trajectories and sample observed traffic states by randomly masking out several intersections.
~\noindent\\~$\bullet$~ \DNY. This is a public traffic that dataset covers a $7\times 28$ network of Upper East Side of Manhattan, New York, collected by both camera data and taxi trajectory data in 2016. This dataset records the lane-level volumes at every 10 second, which means there are 360 time steps in the traffic flow for one hour. We sampled the sparse observations in a similar way as in \DHZ.

\section{Hyperparameter Settings}
\label{app:hyper}
Some of the important hyperparameters can be found in Table~\ref{tab:parameters}.

\begin{table}[h!]
\centering
\caption{Hyper-parameter settings for \ours.}
\label{tab:parameters}

\begin{tabular}{cc | cc}
\toprule
Parameter   & Value & Parameter   & Value \\ \midrule
\begin{tabular}[c]{@{}c@{}}Batch size \end{tabular}                   & 32 &
 Historical time slice $T$            & 30     \\
\begin{tabular}[c]{@{}c@{}}Convolution kernel  \end{tabular}   & 64   &
\begin{tabular}[c]{@{}c@{}}Predict time slice $T'$ \end{tabular}    & 1     \\
\begin{tabular}[c]{@{}c@{}}Time interval of raw data \end{tabular}                & 10(s) &
\begin{tabular}[c]{@{}c@{}}Input channel \end{tabular}                & 11 \\
\begin{tabular}[c]{@{}c@{}}Chebyshev polynomial\end{tabular} & 3    &
\begin{tabular}[c]{@{}c@{}}Output channel\end{tabular} & 3     \\
\begin{tabular}[c]{@{}c@{}}Learning rate\end{tabular}               & 0.001  &
Training epochs         & 100   \\     

\bottomrule
\end{tabular}
\end{table}

\section{Evaluation Metrics}
\label{app:metrics}
We use MAE, RMSE and MAPE to evaluate the performance of predicting next state, which are defined as:
\begin{equation}
    MAE = \frac{1}{N}\frac{1}{F}\mathop{\sum}_{i=1}^{N}\mathop{\sum}_{j=1}^{F}|\obt^{i,j}-\hat{\textbf{x}}^{i,j}_t|
\end{equation}
\begin{equation}
    RMSE = \sqrt{\frac{1}{N}\frac{1}{F}\mathop{\sum}_{i=1}^{N}\mathop{\sum}_{j=1}^{F}||\obt^{i,j}-\hat{\textbf{x}}^{i,j}_t||^2}
\end{equation}
\begin{equation}
    MAPE =\begin{cases}
    & \frac{1}{N}\frac{1}{F}\mathop{\sum}_{i=1}^{N}\mathop{\sum}_{j=1}^{F} \frac{|\obt^{i,j}-\hat{\textbf{x}}^{i,j}_t|}{\obt^{i,j}},\ \obt^{i,j}\not=0\\
    &1, \ \obt^{i,j}=0,\& \obt^{i,j}\not= \hat{\textbf{x}}^{i,j}_t\\
    &0,\ \obt^{i,j}=0,\& \obt^{i,j}=\hat{\textbf{x}}^{i,j}_t \\
\end{cases}
\end{equation}
where $N$ is the total number of road segments, $F$ is the size of observed state vector on the road segment, $\obt^{i,j}$ and $\hat{\textbf{x}}^{i,j}_t$ are the $j$-th element in the traffic volume states of $i$-th road segment at time $t$ in the ground truth and predicted traffic volumes relatively.

\section{Case Study Settings}
\label{app:case}
The traffic signal control task requires the simulator to provide the groundtruth states. After having the states, we mask out the states for several intersections and provide sparse state observations to the traffic signal models. Ideally, the signal control models would take the full states as input and provide actions to the simulator, which would then executes the traffic signal actions from the control method. However, in the sparse data setting, the traffic signal controller at unobserved intersections will not be able to generate any actions since it has no state as input.  With a well-trained state transition model, the full traffic states could be predicted all the intersections, upon which the signal control models would be able to generate actions. 

Following the settings in traffic signal methods~\cite{wei2018intellilight,wei2019presslight},we process the traffic dataset into a format acceptable by the simulator CityFlow~\cite{zhang2019cityflow}. In the processed traffic dataset for the simultor, each vehicle is described as $(o,t,d)$, where $o$ is the origin location, $t$ is time, and $d$ is the destination location. Locations $o$ and $d$ are both locations on the road network. Each green signal is followed by a three-second yellow signal and two-second all red time.

At each time step, we acquire the full state observations from the simulator, mask traffic observations in part of intersections as missing data, and use the state transition model to predict these the next states for all the road segments. Then baseline model ASTGNN and \ours are used as the transition model to recover missing values returned from the simulator and exploit the recovered observations. MaxPressure controller then decides the next traffic signal phase and feed the actions into the simulator. During this process, we can collect groundtruth traffic states from the simulator environment, and predict values from our transition model. After one round of simulation (here we set as 3600 seconds with action interval as 10 seconds), we get the average travel time from the environment and the average queue length for each intersection to evaluate the performance for traffic signal control task. 

\section{Exploration about Loss Function}
\label{app:lossfunc}

We attempt to incorporate graph contrastive learning~\cite{zhu2020deep} into the model to make the model learn additional representations from the prediction. The goal of graph contrastive learning is to learn an embedding space in which positive pairs stay close to each other while negative ones are far apart. In the case of this paper, positive pairs are defined as outputs generated by model using the same input, while negative samples are defined as outputs generated by model using different inputs. In the training process, we first generate
two correlated graph views by randomly performing corruption on the model's output. Then, we use the new loss function to not only minimize prediction loss but also maximize the agreement between outputs in these two views. We select loss function proposed by ~\cite{zhu2020deep} and combine it with Eq.~\eqref{eq:loss} to construct new loss function. In addition, Two loss functions, which combine loss function proposed by ~\cite{zhu2020deep} and its variants separately, are proposed to better adapt to the scenario of traffic networks.
~\noindent\\~$\bullet$~\textbf{$\mathcal{L}_N$}. For any observable node $i$, its two outputs generated by model at time $T+1$, $u_{T+1}^i$ and $v_{T+1}^i$, form the positive sample, and outputs of nodes other than $i$ in the two views are regarded as negative samples. Formally, we define the critic $d(u,v)=s(f(u),f(v))$, where $s$ is the cosine similarity and $f$ is the output of model. The pairwise objective for each positive pair ($u_{T+1}^i,v_{T+1}^i$) is defined as:
\begin{align}
\begin{split}\label{eq:dis}
    \ell (u_{T+1}^i,v_{T+1}^i)=\log \frac{e^{d(u_{T+1}^i,v_{T+1}^i)/\tau}}{e^{d(u_{T+1}^i,v_{T+1}^i)/\tau}+\sum _{k=1}^{N}\mathbbm{1}_{[k\ne i]}e^{d(u_{T+1}^i,v_{T+1}^i)/\tau }}
\end{split}
\end{align}

Where $\mathbbm{1}_{k\ne i} \in\{0,1\}$ is an indication that equals to 1 if $k\ne i$, and $\tau$ is a temperature parameter. Since two views are symmetric, the loss at $T+1$ for another view is defined similarly for $\ell (v_{T+1}^i,u_{T+1}^i)$. The definition of $\mathcal{L}_N$ is as follows: 
\begin{align}
\begin{split}\label{eq:lossn}
    \mathcal{L}_N=\mathcal{L}_p+\frac{1}{2N}\sum_{i=1}^{N}[\ell (u_{T+1}^i,v_{T+1}^i)+\ell (v_{T+1}^i,u_{T+1}^i)]  
\end{split}
\end{align}

Where $\mathcal{L}_p$ represent loss function of Eq.~\eqref{eq:loss}. The goal of the remainder of $\mathcal{L}_N$ is to minimize the distance between pairs of positive samples and maximize the distance between pairs of negative samples.
~\noindent\\~$\bullet$~\textbf{$\mathcal{L}_{N-N}$}. Based on $\mathcal{L}_N$, $\mathcal{L}_{N-N}$ removes denominator of Eq.~\eqref{eq:dis}. It can be considered that we only need to make the positive pairs as close as possible regardless of the distance between the negative pairs. It is reasonable in traffic networks, for road segments that are considered main roads may have similar traffic flow values at the same time interval. The definition of $\mathcal{L}_{N-N}$ is as follows:
\begin{align}
\begin{split}\label{eq:lossn}
    \mathcal{L}_{N-N}=\mathcal{L}_p+\frac{1}{2N}\sum_{i=1}^{N}[\ell ^- (u_{T+1}^i,v_{T+1}^i)+\ell ^- (v_{T+1}^i,u_{T+1}^i)]
\end{split}
\end{align}

Where $\ell ^- (u_{T+1}^i,v_{T+1}^i)=d(u_{T+1}^i,v_{T+1}^i)/\tau$.

Table~\ref{tab:lossfunc} shows the performance of the loss function and its variants. We can observe that \ours with $\mathcal{L}_{N-N}$ outperforms \ours with $\mathcal{L}_N$, indicating that there is no obvious negative pairs relationship between outputs produced by different inputs under the traffic scenario. Moreover, \ours further outperforms \ours with $\mathcal{L}_{N-N}$. We believe that the model's method of constructing positive pairs may be too arbitrary to learn additional information, resulting in suboptimal performance.

\begin{table}[h!]
\centering
\caption{Performance of \ours using different loss functions on \DHZ. The lower, the better. Similar trends are also found on other datasets.}
\begin{tabular}{cccc}
\toprule
Method & MSE    & RMSE   & MAPE   \\ \midrule
DTIGNN with $\mathcal{L}_N$  & 0.4588 & 0.8359 & 0.2969 \\
DTIGNN with $\mathcal{L}_{N-N}$  & 0.4355 & 0.7282 & 0.2876 \\ \midrule
Ours   & 0.3810  & 0.6618 & 0.2455 \\
\bottomrule
\end{tabular}
\label{tab:lossfunc}
\end{table}

\end{sloppypar}
\end{document}